\documentclass[10pt,twocolumn,letterpaper]{article}

\usepackage{iccv}
\usepackage{times}
\usepackage{epsfig}
\usepackage{graphicx}
\usepackage[caption=false,font=footnotesize]{subfig}
\usepackage{amsmath}
\usepackage{amssymb}
\usepackage{enumitem}
\usepackage[pagebackref=true,breaklinks=true,letterpaper=true,colorlinks,bookmarks=false]{hyperref}
\usepackage{mathrsfs}
\usepackage[english]{babel}
\usepackage{blindtext}
\usepackage{mathtools}
\usepackage{bm}
\usepackage{multirow}

\iccvfinalcopy
\def\iccvPaperID{68}

\newcommand*{\affaddr}[1]{#1} 
\newcommand*{\affmark}[1][*]{\textsuperscript{#1}}
\newcommand*{\email}[1]{\texttt{#1}}

\ificcvfinal\fi
\begin{document}

\title{Multi-View Image Generation from a Single-View}


\author{%
Bo Zhao\affmark[1,2] \qquad Xiao Wu\affmark[1] \qquad Zhi-Qi Cheng\affmark[1] \qquad Hao Liu\affmark[2] \qquad Zequn Jie\affmark[3]\qquad Jiashi Feng\affmark[2]\\
\affaddr{\normalsize \affmark[1]Southwest Jiaotong University} \quad
\affaddr{\affmark[2]National University of Singapore} \quad
\affaddr{\affmark[3]Tencent AI Lab}\\
\email{\small {\{zhaobo.cs, wuxiaohk, zhiqicheng, hfut.haoliu, zequn.nus\}@gmail.com}, elefjia@nus.edu.sg}
}

\maketitle

\begin{abstract}
This paper addresses a challenging problem -- how to generate multi-view cloth images from only a single view input. To generate realistic-looking images with different views from the input, we propose a new image generation model termed VariGANs that combines the strengths of the variational inference and the Generative Adversarial Networks (GANs). Our proposed VariGANs model generates the target image in a coarse-to-fine manner instead of a single pass which suffers from severe artifacts. It first performs variational inference to model global appearance of the object (e.g., shape and color) and produce a coarse image with a different view. Conditioned on the generated low resolution images, it then proceeds to perform adversarial learning  to fill details and generate images of consistent details with the input. Extensive experiments conducted on two clothing datasets, MVC and DeepFashion, have demonstrated that images of a novel view generated by our model are more plausible than those generated by existing approaches, in terms of more consistent global appearance as well as richer and sharper details.
\end{abstract} 

\section{Introduction}


Products at e-commerce websites are usually displayed by images from different views. Multi-view images provide straightforward and comprehensive product illustrations to potential buyers. However, such multi-view images are often expensive to produce in both time and cost, thus sometimes not available. 
For example, when one occasionally sees an image of the desired clothing item from a magazine, which only provides a single view image, he/she has to imagine its look from other views. An automatic model that can generate multi-view images from a single-view input is desired in such scenarios and can find practical application on e-commerce platforms and other applications like photo/video editing and AR/VR. 
Provided a single view clothing image, we aim to generate the rest views of the input image without requiring any extra information. 

\begin{figure}[t!]
  \centering
  \includegraphics[width=0.9\columnwidth]{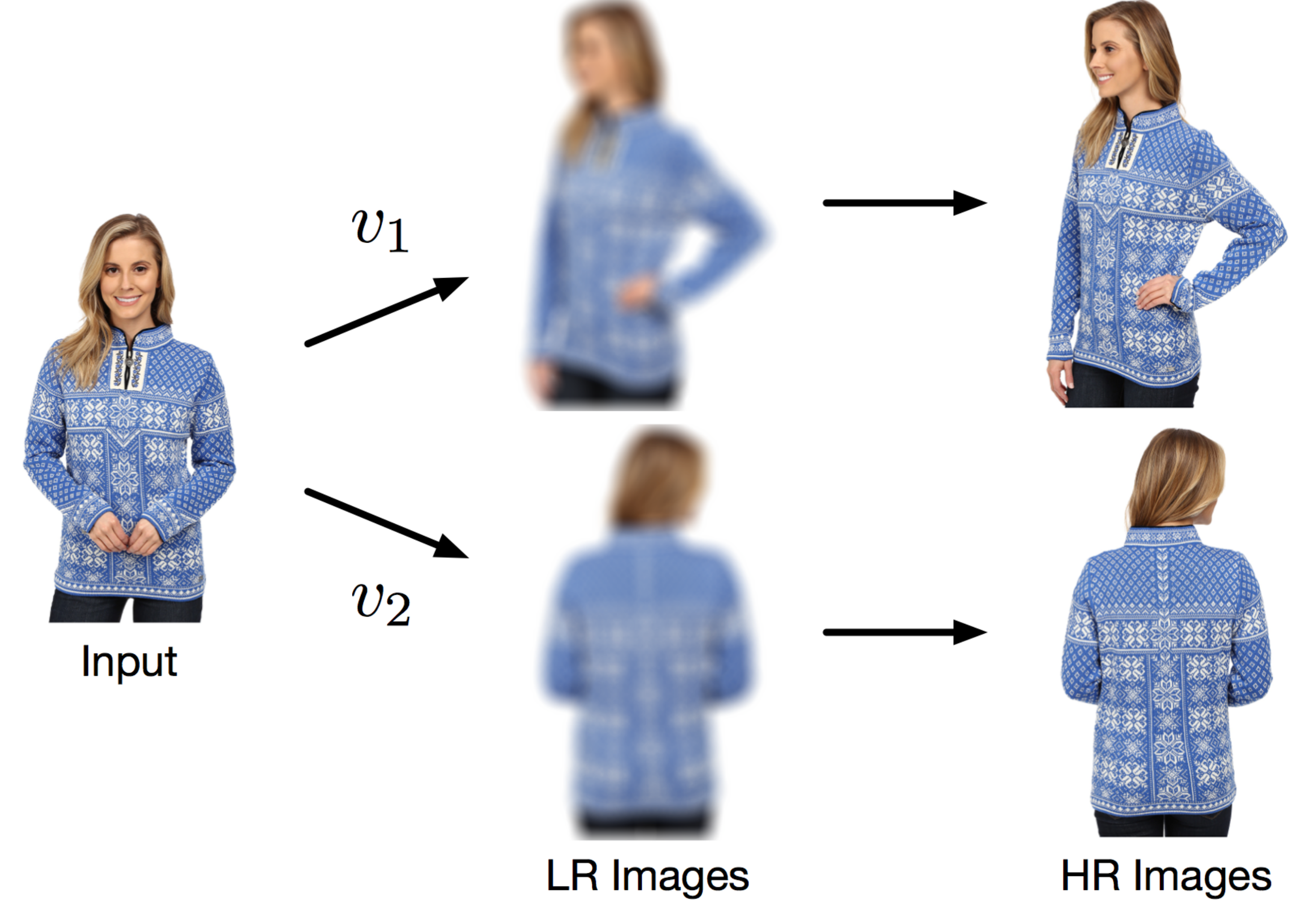}
  \caption{The photo-realistic image generation process of the proposed VariGANs. It decomposes the image generation into low and high image generation. The LR images are firstly generated by variational inference with new views $v_1$ and $v_2$. Then, the HR images are generated by filling the details and correcting the defects through adversarial learning.}
  \label{fig: examples}
  \vspace{-0.2in}
\end{figure}

Although image generation is a challenging task due to the high dimensionality of images and the complex configuration and layout of image contents, some recent works have demonstrated good performance on realistic image generation beneficial from advanced models like Variational Autoencoder (VAE)~\cite{Kingma2014}) and Generative Adversarial Networks (GANs)~\cite{Goodfellow2014}. 
VAE adopts variational inference plus deep representation learning to learn a complex generative model and gets rid of the time-consuming sampling process.  However, VAE usually fails to provide rich details in generated images.  Another popular generative model, GANs, introduces a real-fake discriminator to supervise the learning of the generative model. Benefiting from the competition between discriminator and generator, GANs are advantageous in providing realistic details, but they usually introduce artifacts to the global appearance, especially when the image to be generated is large. 

To tackle this challenging problem of generating multi-view images from a single-view observation, many approaches~\cite{Chen2013, Kholgade2014, Zheng2012} first construct the 3D structure of the object and then generate desired target view images from that model. While other methods~\cite{Park2017, Yan2016a, Zhou2016a} learn the transformation between the input view and target view by relocating pixels. However, those methods mainly synthesize rigid objects, \eg cars, chairs with simple textures. The generation of deformable objects with rich details such as clothes or human body has not been fully explored.  

In this paper, we propose a novel image generation model named Variational GAN (VariGAN) that combines the strengths of variational inference and adversarial training. The proposed model overcomes the limitations of GAN in modeling global appearance by introducing internal variational inference in the generative model learning. A low resolution (LR) image capturing global appearance is firstly generated by variational inference. This process learns to draw rough shapes and colors of the images to generate at a different view,  conditioned on the given images. With the generated LR image, VariGAN then performs adversarial learning to generate realistic high resolution (HR) images by filling richer details to the low resolution image. Since the LR image only has the basic contour of the target object in a desired view, the fine image generation module just needs to focus on drawing details and rectifying defects in low resolution images. See Fig.~\ref{fig: examples} for illustration. Decomposing the complicated image generation process into the above two complementary learning processes significantly simplifies the learning and produces more realistic-look multi-view images. 
Note that VariGAN is a generic model and can be applied to other image generation applications like style transfer. We would like to exploit these potential applications of VariGAN in the future. 

The main contributions are summarized as follows:

\begin{enumerate}[label={(\arabic*)}]
  \item To our best knowledge, we are the first to address the new problem of generating multi-view clothing images based on a given clothing image of a certain view, which has both theoretical and practical significance. 
  \item We propose a novel VariGAN generation architecture for multi-view clothing image generation that adopts a new coarse-to-fine image generation strategy. The proposed model is effective in both capturing global appearance and drawing richer details consistent with the input conditioned image. 

  \item We apply our proposed model on two largest clothes image datasets and demonstrate its superiority  through comprehensive evaluations compared with other state-of-the-arts. We will release the model and relevant code upon acceptance. 
\end{enumerate}

\section{Related Work}
\paragraph{Image Generation.}
Image generation has been a heated topic in recent years. Many approaches have been proposed with the emergence of deep learning techniques. Variational Autoencoder (VAE)~\cite{Kingma2014} generates images based on the probabilistic graphical models, and are optimized by maximizing the lower bound of the data likelihood. Yan~\etal~\cite{Yan2016} proposed the Attribute2Image, which generates images from visual attributes. They modeled an image as a composite of foreground and background and extended the VAE with disentangled latent variables. Gregor~\etal~\cite{Gregor2015} proposed the DRAW, which integrates the attention mechanism to the VAE to generate realistic images recurrently by patches. Different from the generative parametric approaches, Generative Adversarial Networks (GANs) proposed by Goodfellow~\etal~\cite{Goodfellow2014} introduce a generator and a discriminator in their model. The generator is trained to generate images to confuse the discriminator, and the discriminator is trained to distinguish between real and fake samples. Since then, many GANs-based models have been proposed, including Conditional GANs~\cite{Mirza2014}, BiGANs~\cite{Donahue2016, Dumoulin2017}, Semi-suprvised GANs~\cite{Odena2016}, InfoGAns~\cite{Chen2016} and Auxiliary Classifier GANs~\cite{Odena2016a}. GANs have been used to generate images from labels~\cite{Mirza2014}, text~\cite{Reed2016, Zhang2016} and also images~\cite{Isola2016, Pathak2016, Salimans2016, Yoo2016, Zhu2016, Zhou2016}. Our proposed model is also an image-conditioned GAN, with generation capability strengthened by variational inference. 

\vspace{-0.1in}
\paragraph{View Synthesizing.}
Images with different views of the object can be easily genertaed with the 3D modeling of the object~\cite{Chen2013, Dosovitskiy2015, Kulkarni2015, Kholgade2014, Zheng2012}. Hinton~\etal~\cite{Hinton2011} proposed a transforming auto-encoder to generate images with view variance. Rezende~\etal~\cite{Rezende2016} introduced a general framework to learn 3D structures from 2D observations with a 3D-2D projection mechanism. Yan~\etal~\cite{Yan2016a} proposed Perspective Transformer Nets to learn the projection transformation after reconstructing the 3D volume of the object. Wu~\etal~\cite{Wu2016a} also proposed the 3D-2D projection layers that enable the learning of 3D object structures using annotated 2D keypoints. They further proposed the 3D-GAN~\cite{Wu2016a} which generates 3D objects from a probabilistic space by leveraging recent advances in volumetric convolutional networks and generative adversarial nets. Zhou~\etal~\cite{Zhou2016a} propose to synthesize novel views of the same object or scene  corresponding by learning appearance flows. Most of these models are trained with the target view images or image pairs which can be generated by a graphics engine. Therefore, in theory, there are infinite amount of training data with desired view to train the model. However, in our task, the training data is limited in both views and numbers, which greatly adds the difficulties to generate image of different views. 

\section{Proposed Method}

\subsection{Problem Setup}

\begin{figure*}[!t]
  \centering
  \includegraphics[width=1\hsize]{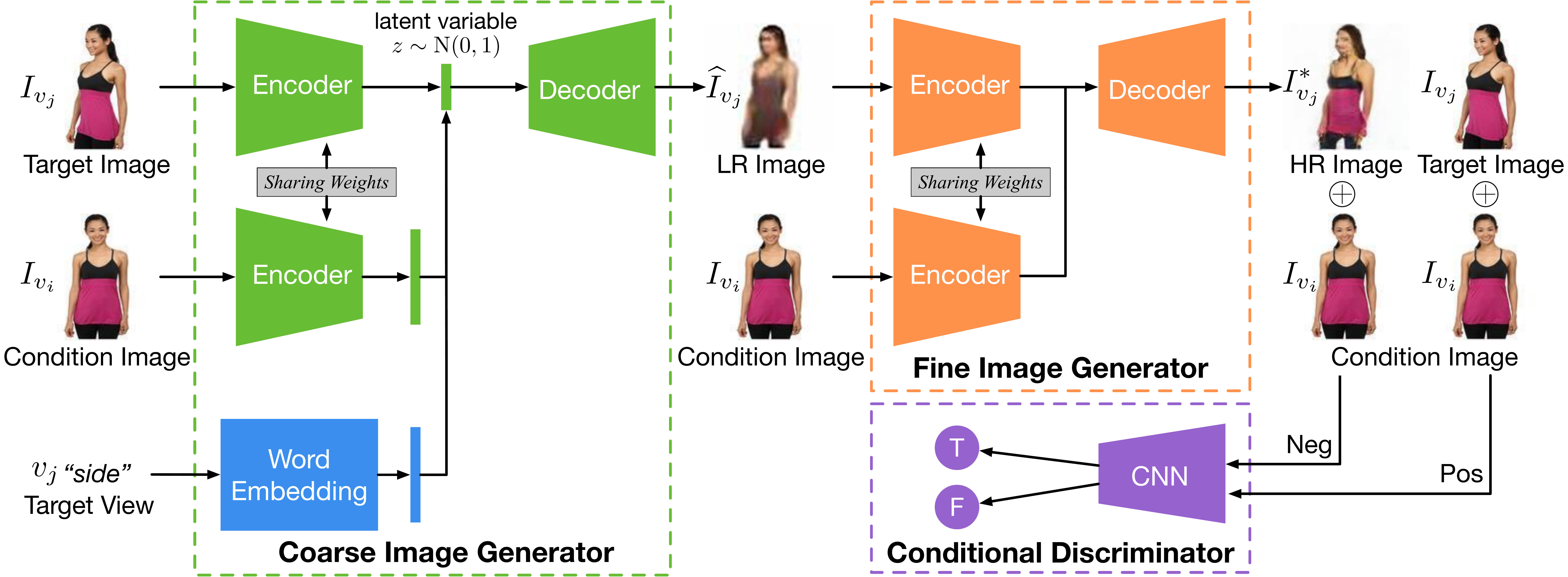}
  \caption{Architecture of the proposed VariGAN. It consists of three modules: coarse image generator, fine image generator and  conditional discriminator. During training, a LR Image is firstly generated by the coarse image generator conditioned on the target image, conditioned image and target view. The fine image generator with skip connections is designed to generate the HR image. Finally, the HR image and the conditioned image are concatenated as negative pairs and passed to the conditional discriminator together with positive pairs (target image \& condition image) to distinguish real and fake.}
  \label{fig: arch}
  \vspace{-0.2in}
\end{figure*}

We first define the problem of generating multi-view images from a single view input. Suppose we have a pre-defined set of view angles $\mathbf{V}=\{v_1, \cdots, v_i, \cdots, v_n\}$, where $v_i$ corresponds to one specific view, \eg front or side view. An object captured from the view $v_i$ is denoted as $I_{v_i}$. Given the source image $I_{v_i}$, multi-view image generation is to generate an image $I_{v_j}$ with a different view from $I_{v_i}$, \ie $v_j \in \mathbf{V}$ and $j \neq i$. 


\subsection{Variational GANs}
 
 
Standard GANs have been applied to generate images of desired properties  based on  the input.
This type of model learns a generative  model $G$ on the distribution  of the desired images, sampling from which would provide new images.  Different from other generative models, GANs employ an extra discriminative model $D$ to supervise the generative learning process, instead of purely optimizing $G$ to best explain training data via Maximum Likelihood Estimation.

Specified in multi-view image generation, given an input conditioned image $I_{v_i}$ (captured at viewpoint $v_i$), the goal is  to learn the distribution $p(I_{v_j}|I_{v},v_j)$ for generating the new image $I_{v_j}$ of a different viewpoint $v_j$, from a labeled dataset $(I_{v_j,1}, I_{v_i,1}), \ldots, (I_{v_j,n}, I_{v_i,n})$. Here $v_j$ is specified by users as an input to the model.
 
The objective of GANs is defined as
\begin{align}
\vspace{-0.2in}
  \min_{\theta_G} \max_{\theta_D}& \mathbb{E}_{{I_{v_i}\sim p_{\text{data}}(I_{v_i}), I_{v_j}\sim p_{\text{data}}(I_{v_j}|I_{v_i}, v_j)}}[\log D(I_{v_i}, I_{v_j})]+\nonumber\\
  &\mathbb{E}_{z\sim p(z)}[\log(1-D(I_{v_i}, G(z, I_{v_i}, v_j))],\nonumber
\vspace{-0.2in}
\end{align}
where $G$ tries to generate real data $I_{v_j}$ given noise $z \sim p(z)$ through  minimizing its loss  to fool  an adversarial discriminator $D$, and $D$ tries to maximize its discrimination accuracy between real data and generated data.
  
However, it is difficult to learn a generator  $G$ to produce  plausible images of  high resolution, correct contour and rich details, because GANs are limited in capturing global appearance. To address this critical issue and generate more realistic images, the variational GANs (VariGANs) proposed in this work combines the strengths of variation inference for modeling correct contours  and adversarial learning to fill realistic details. It decomposes the generator into two components. One is for generating a coarse image through the variational inference model $V$ and the other is for generating the final image with fine details based on the outcome from $V$. Formally, the objective of  VariGANs is
\begin{align}
\vspace{-0.2in}
\label{eq: VariGANs_obj}
&\min_{\theta_G}\max_{\theta_D, \theta_V} \mathbb{E}_{{I_{v_i}\sim p_{\text{data}}(I_{v_i}), I_{v_j}\sim p_{\text{data}}(I_{v_j}|I_v, v_j)}}[\log D(I_{v_i}, I_{v_j})]+\nonumber\\
&\quad\mathbb{E}_{z\sim p(z)}[\log(1-D(I_{v_i}, G(V(z, I_{v_i}, v_j), I_{v_i},  v_j))].
\vspace{-0.2in}
\end{align}
Here $z$ is the random latent variable and $V$ is the coarse image generator. This objective can be optimized by maximizing the variational lower bound of $V$, maximizing the discrimination accuracy of $D$, and minimizing the loss of $G$ against $D$. 
We will elaborate the model of $V$, $G$ and $D$ in the following parts.

\subsection{Coarse Image Generation}

Given an input image $I_{v_i}$ with the  view  of $v_i$, target view $v_j$, and latent variable $z$, the coarse image generator $V(I_{v_i}, z, v_j)$ learns the distribution  $p(\widehat{I}_{v_j}|I_{v}, z)$ with focus on modeling the global appearance. We use $\theta_V$ to denote parameters of the coarse image generator. To alleviate difficulties of directly optimizing this log-likelihood function and avoid the time-consuming sampling, we apply the variational Bayesian approach to optimize the lower bound of the log-likelihood $\log p_\theta(\widehat{I}_{v_j}|I_{v}, v_j)$, as proposed in \cite{Kingma2014, Rezende2014}. Specifically, an auxiliary distribution $q_\phi(z|\widehat{I}_{v_j}, I_{v_i}, v_j)$ is introduced to approximate the true posterior $p_{\theta_V}(z|\widehat{I}_{v_j},I_{v_i}, v_j)$.
 
The conditional log-likelihood of the coarse image generator $V$ is defined as
\begin{align}
\vspace{-0.2in}
   & \log p_{\theta_V}(\widehat{I}_{v_j}|I_{v_i}, v_j) =  \mathcal{L}(\widehat{I}_{v_j},I_{v_i},v_j;\theta, \phi) + \nonumber \\ 
    & \quad \qquad \mathrm{KL}\left(q_\phi(z|\widehat{I}_{v_j},I_{v_i},v_j)||p_\theta(z|\widehat{I}_{v_j},I_{v_i},v_j)\right), \nonumber
\vspace{-0.2in}
\end{align}
where the variational lower bound is
\begin{align}
\vspace{-0.2in}
  \mathcal{L}(\widehat{I}_{v_j},&I_{v_i},v_j;\theta, \phi)=-\mathrm{KL}\left(q_\phi(z|\widehat{I}_{v_j},I_{v_i},v_j)||p_\theta(z)\right) \nonumber\\
&+{E}_{q_\phi(z|\widehat{I}_{v_j},I_{v_i},v_j)}[\log p_\theta(\widehat{I}_{v_j}|I_{v_i},v_j,z)],
  \label{eq: lower-bound}
\vspace{-0.2in}
\end{align}
where the first $\mathrm{KL}$ term in Eqn.~\eqref{eq: lower-bound} is a regularization term that reduces the gap between the prior $p(z)$ and the proposal distribution $q_\phi(z|\widehat{I}_{v_j},I_{v_i},v_j)$. The second term $\log p_{\theta_V}(\widehat{I}_{v_j}|I_{v_i},v_j, z)$ is the log likelihood of samples and is usually measured by the reconstruction loss, \eg, $\ell_1$ used in our model.

\subsection{Fine Image Generation}
After obtaining the low resolution image $\widehat{I}_{v_j}$ of the desired output  $I_{v_j}$, the fine image generation module learns another generator $G$ that maps the low resolution image $\widehat{I}_{v_j}$ to the high resolution image $I_{v_j}^*$ conditioned on the input $I_{v_i}$. The generator $G$ is trained to generate images that cannot be distinguished from ``real'' images by an adversarial conditional discriminator, $D$, which is trained to distinguish as well as possible the generator's ``fakes''. See Eqn.~\eqref{eq: VariGANs_obj}.
 
Since the multi-view image generator need not only fool the discriminator but also be near the ground truth of the target image with a different view, we also add the $\ell_1$ loss for the generator. The $\ell_1$ loss is chosen because it alleviates over-smoothing artifacts compared with $\ell_2$ loss.
 
Then, the GANs of fine image generation train the discriminator $D$ and the generator $G$ by alternatively maximizing $\mathcal{L}_D$ in Eqn.~\eqref{eq: gan_ld} and minimizing $\mathcal{L}_{G}$ in Eqn.~\eqref{eq: gan_lg}:
\begin{align}
  \mathcal{L}_D=&\mathbb{E}_{(I_{v_i}, I_{v_j})\sim p_{\text{data}}}[\log D(I_{v_i}, I_{v_j})]+\nonumber\\
  &\mathbb{E}_{z \sim p(z) } [\log (1-D(I_{v_i}, G(\widehat{I}_{v_j}(z), I_{v_i},v_j)))] \label{eq: gan_ld},\\
  \mathcal{L}_G=&\mathbb{E}_{z \sim p(z) }[\log (1-D(I_{v_i}, G(\widehat{I}_{v_j}(z), I_{v},v^\prime)]+\nonumber\\
  &\lambda \| I_{v_j} - G(\widehat{I}_{v_j}(z), I_{v_i},v_j) \|_1,\label{eq: gan_lg}
\end{align}
where $\widehat{I}_{v_j}$ is the coarse image generated by $V$. The real images $I_{v_i}$ and $I_{v_j}$ are from the true data distribution.

\subsection{Network Architecture}

The overall architecture of the proposed model in the training phase is illustrated in Fig.~\ref{fig: arch}. It consists of three modules: the coarse image generator, the fine image generator and the conditional discriminator. 
During training, the target view image $I_{v_j}$ and the conditioned image $I_{v_i}$ are passed through two siamese-like encoders to learn their representations respectively. By word embedding, the input desired view angle $v_j$ is transformed into a vector. The representations of $I_{v_i}$, $I_{v_j}$ and $v_j$ are combined to generate the latent variable $z$. 
However, during testing, there is no target image $I_{v_i}$ and the encoder for it. The latent variable $z$ is randomly sampled and combined with the representation of the condition image $I_{v_i}$ and the target view $v_j$ to generate the target view LR image $\widehat{I}_{v_j}$.
After that, $I_{v_i}$ and $\widehat{I}_{v_j}$ are sent to the fine image generator to generate the HR image $I_{v_j}^*$. Similar to the coarse image generation module, the fine image generation module also contains two siamese-like encoders and a decoder. Moreover, there are skip connections between mirrored layers in the encoder and decoder stacks. 
By the channel concatenation of the HR image $I_{v_j}^*$ and the condition image $I_{v_i}$, a conditional discriminator is adopted to distinguish whether the generated image is real or fake.

\vspace{-0.15in}
\paragraph{Coarse Image Generator}
There are several convolution layers in the encoder of the coarse image generator to down sample the input image to an $M_l \times 1 \times 1$ tensor. Then a fully-connected layer is topped to transform the tensor to an $M_l$-D representation. The encoders for the target image and the condition image share weights. 
A word embedding layer is employed to embed the target view into an $M_l$-D vector. The representations of the target image, the conditioned image and the view embedding are combined and transformed to an $M_l$-D latent variable.
Then, the latent variable together with the conditioned image representation and the view embedding are passed through a series of de-convolutional layers to generate a $W_{\mathrm{LR}} \times W_{\mathrm{LR}}$ image.

\vspace{-0.15in}
\paragraph{Fine Image Generator with Skips}
Similar to the coarse image generation module, the fine image generator also contains two siamese-like encoders and a decoder. The encoder consists of several convolutional layers to down-sample the image to a $M_h \times 1 \times 1$ tensor. Then several de-convolutional layers are used to up-sample the bottleneck tensor to $W_{\mathrm{HR}}\times W_{\mathrm{HR}}$.

\begin{figure}[!t]
  \centering
  \includegraphics[width=0.8\columnwidth]{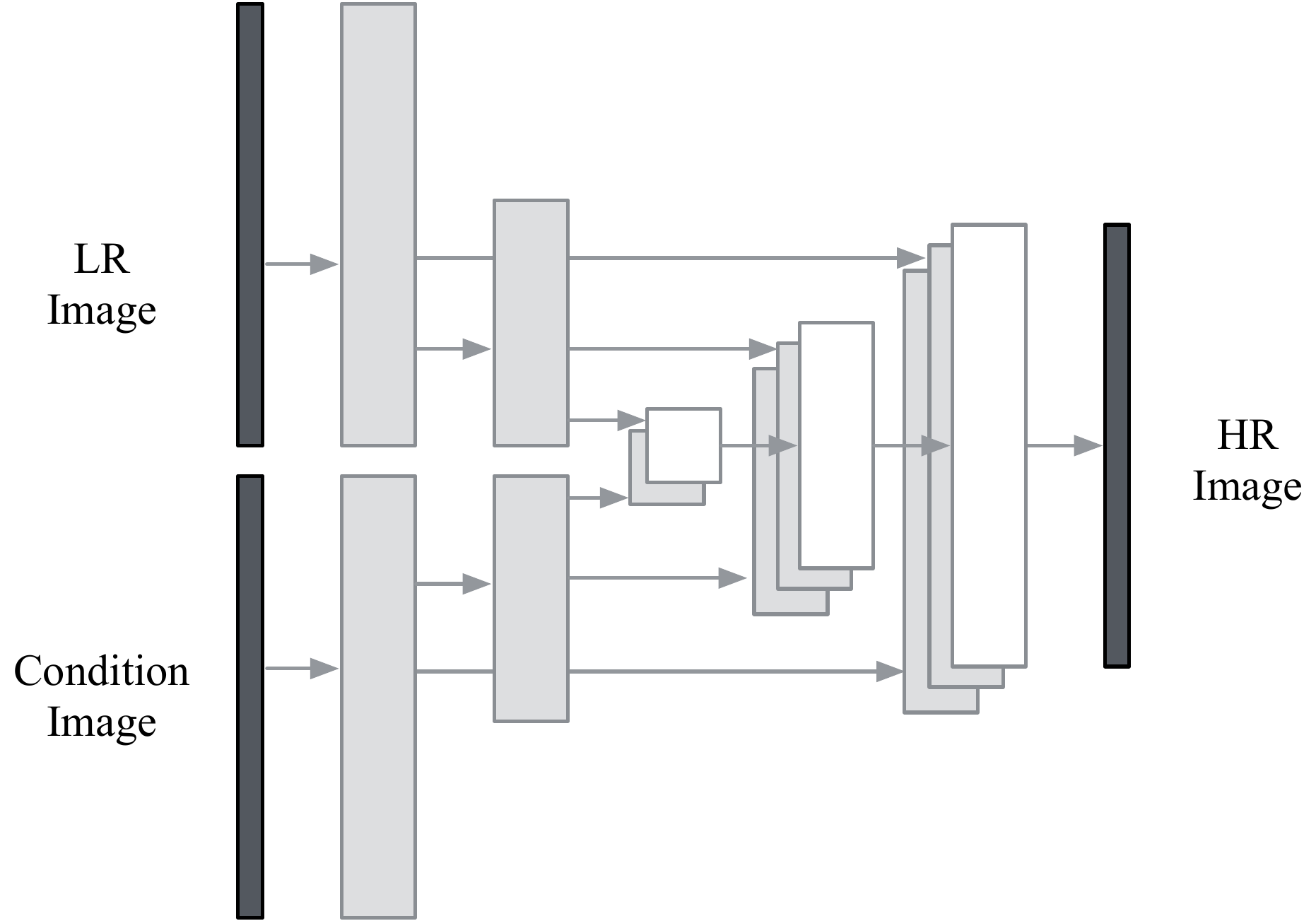}
  \caption{Dual-path U-Net. There are skip connections between the mirrored layers in two encoders and a decoder.}
  \label{fig-y-net}
  \vspace{-0.2in}
\end{figure}

Since the mapping from low resolution image to high resolution image can be seen as a conditional image translation problem, the low and high resolution images only differ in surface appearance, but both are rendered under the same underlying structure. Therefore, the shape information can be shared between the LR and HR image. Besides, the low-level information of the conditioned image will also provide rich guidance when translating the LR image to the HR image. It would be desirable to shuttle these two kinds of information directly across the net. Inspired by the work of ``U-Net''~\cite{Ronneberger2015} and image-to-image translateion~\cite{Isola2016}, we add skip connections between the LR image encoder and the HR image decoder, and also between the conditioned image encoder and the HR image decoder  simultaneously (see Fig.~\ref{fig-y-net}). 
By such skip connections, the decoder up-samples the encoded tensor to the high resolution image with the target view by several de-convolution layers.

\vspace{-0.1in}
\paragraph{Conditional Discriminator} 
The generated high resolution image $I_{v_j}^*$ and the ground-truth target image $I_{v_j}$ are concatenated with the conditioned image $I_{v_i}$ by channels to form the negative pair and positive pair, respectively. These two kinds of image pairs are passed to the conditional discriminator and train the fine image generator adversarially.

%
%
%


\subsection{Implementation Details}

For the coarse image generator, an encoder network contains 6 convolution layers followed by 1 fully-connected layer (convolution layers have 64, 128, 256, 256, 256 and 1024 channels with filter size of 5$\times$5, 5$\times$5, 5$\times$5, 3$\times$3, 3$\times$3 and 4$\times$4, respectively; the fully-connected layer has 1024 neurons), and $M_l$ is set to 1024, $W_{\mathrm{LR}}$ is set to 64, respectively. The representations of the target image and the condition image and the embedding of the target view are concatenated and transformed to the latent variable by a fully-connected layer with 1024 neurons. The decoder network consists of 1 fully-connected layers with 256$\times$8$\times$8 neurons, followed by 6 de-convolutional layers with 2$\times$2 up-sampling (with 256, 256, 256, 128, 64 and 3 channels with filter size of 3$\times$3, 5$\times$5, 5$\times$5, 5$\times$5, 5$\times$5 and 5$\times$5). 

For the fine image generation module, the encoder network contains 7 convolution layers (with 64, 128, 256, 512, 512, 512, 512 channels with filter size of 4$\times$4 and stride 2). Thus, $M_h$ is set to 512. The decoder network consists of 7 de-convolutional layers with 512, 512, 512, 256, 128, 64 and 3 channels with filter size 4$\times$4 and stride 2. The conditional discriminator consists of 5 convolutional layers (they have 64, 128, 256, 512 and 1 channel(s) with filer size 4$\times$4 and stride 2, 2, 2, 1, 1). $W_{\mathrm{HR}}$ is set to 128.

For training, we first train the coarse image generator for 500 epochs. Using the generated low resolution image and the conditioned image, we then iteratively train the fine image generator and the conditional discriminator for another 500 epochs. All networks are trained using ADAM solvers with batch size of 32 and an initial learning rate of 0.0003.

\section{Experiment}

To validate the effectiveness of our proposed VariGAN model, we conduct extensive quantitative and qualitative evaluations on the MVC~\cite{Liu2016} and the DeepFashion~\cite{Liu2016a} datasets that contain a huge number of clothing images with different views. We compare the performance of generating multi-view images with two state-of-the-art image generation models:~conditional VAE~(cVAE)~\cite{Sohn2015},~conditional GANs~(cGANs)~\cite{Mirza2014}. 

The cVAE has a similar architecture as the coarse image generator in our VGANs. It has one more convolution layer in the encoder and one more de-convolution layer in the decoder, which directly generates the HR Image. The cGANs have one encoder network to encode the conditioned image and one word embedding layer to transform the view to the vector. The encoded conditioned image and the view embedding are concatenated and fed into the decoder to generate the HR image.  

In addition to performance comparison with state-of-the-art models, we do ablation studies to investigate the design and important components of our proposed VariGANs. We firstly conduct experiment that replace the variational inference with GANs. Secondly, we train our model without the dual-path U-Net in the fine image generation module to verify the role of the skip connections between the encoders and the decoder. We also conduct experiments without $\ell_1$ loss to prove the importance of the traditional loss for plausible image generation. Finally, we do experiments of VariGANs without the conditional discriminator to show whether the channel-concatenation of the generated image and the condition image is beneficial for the high resolution image generation. Now we proceed to introduce details on evaluation benchmarks and present experimental results. 

\subsection{Datasets and Evaluation Metrics}

\begin{figure}[t!]
  \centering
  \subfloat[MVC]{\includegraphics[width=0.48\columnwidth]{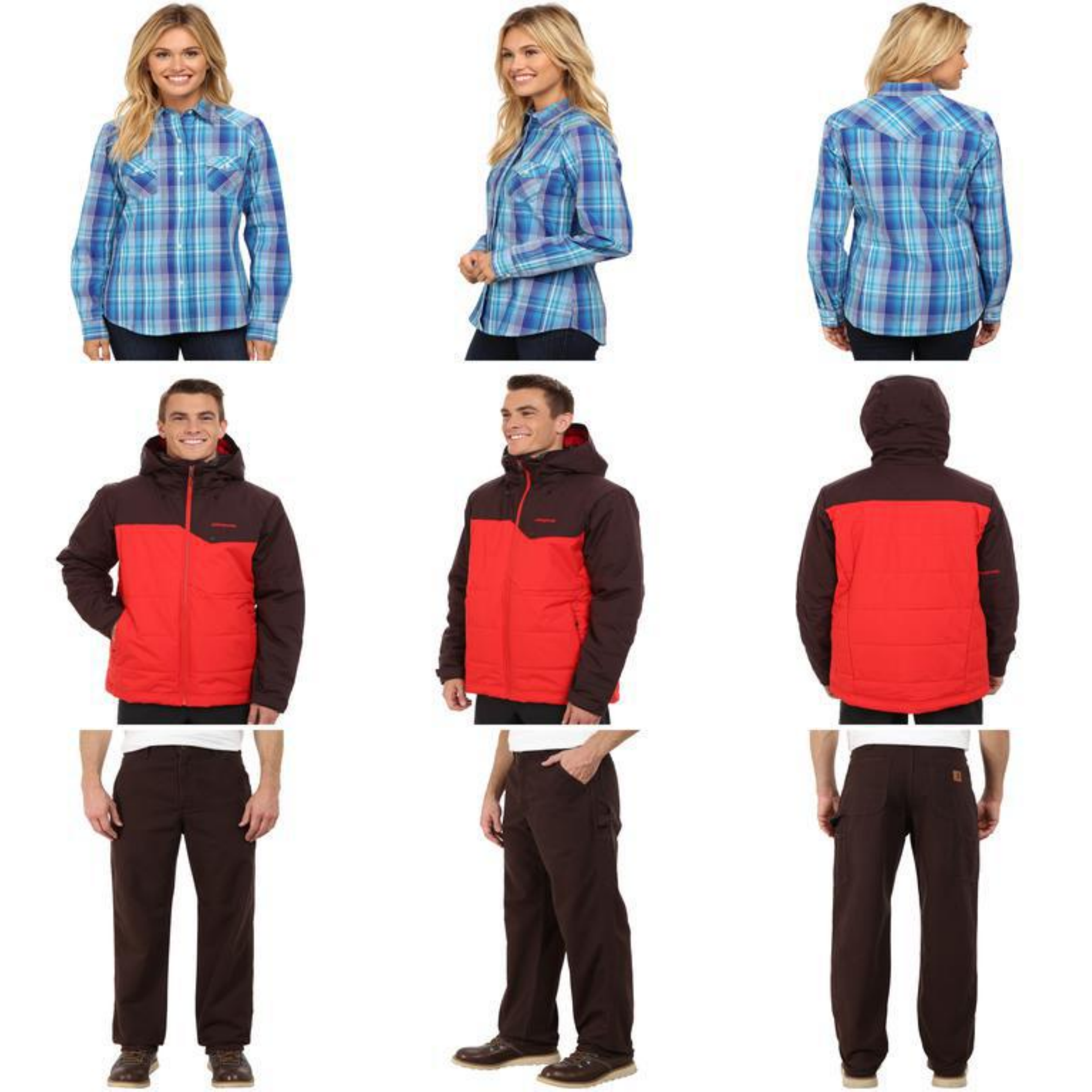}}
  \hspace{0.02in}
  \subfloat[DeepFashion]{\includegraphics[width=0.48\columnwidth]{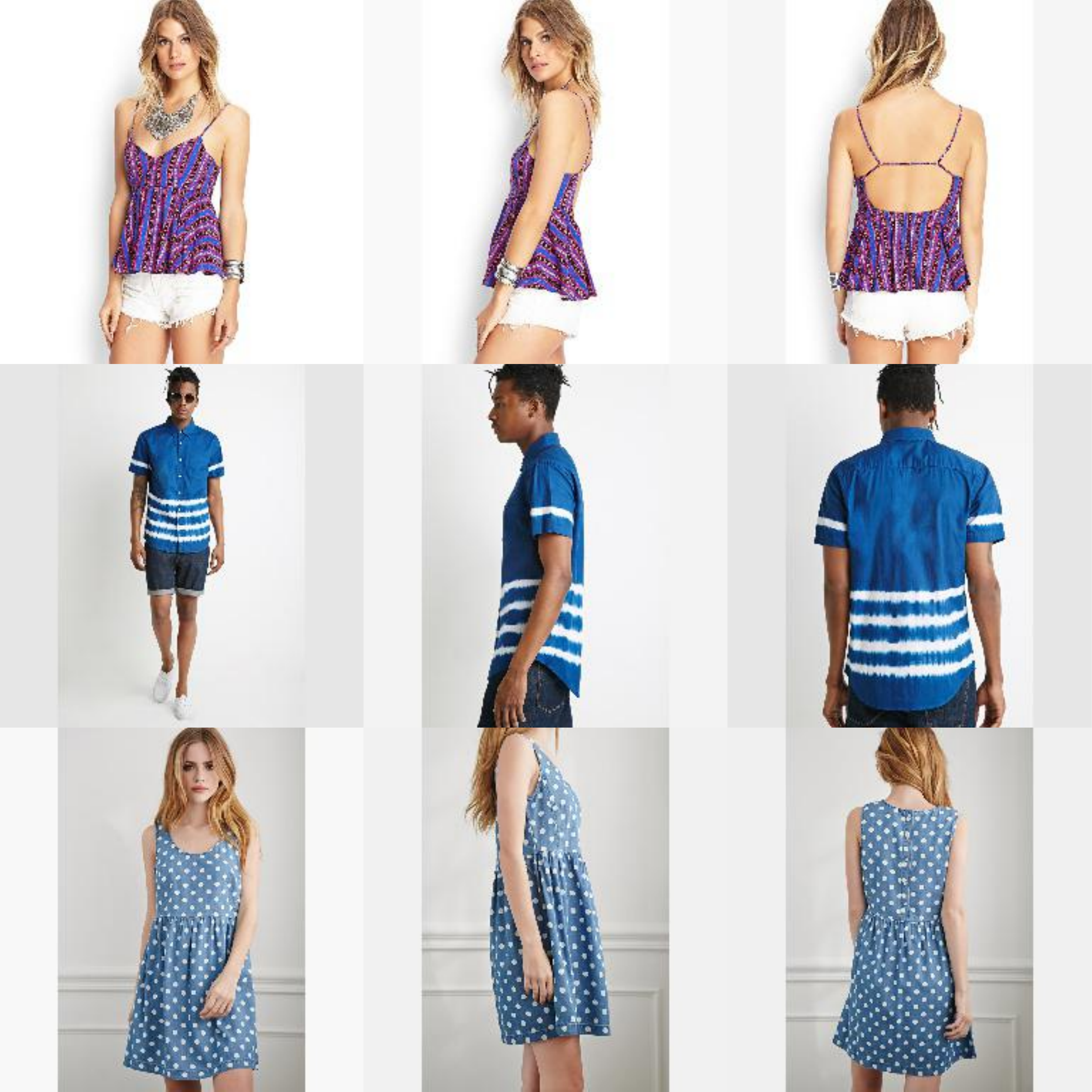}}
  \caption{Example images with multi-views from the MVC~\cite{Liu2016} and DeepFashion~\cite{Liu2016a}, respectively.}
  \label{fig: problem-setup}
  \vspace{-0.2in}
\end{figure}

\paragraph{Datasets} MVC~\footnote{http://mvc-datasets.github.io} contains 36,323 clothing items. Most of them have 4 views, \ie front, left side, right side and back side. DeepFashion~\footnote{http://mmlab.ie.cuhk.edu.hk/projects/DeepFashion.html} contains 8,697 clothing items with 4 views, \ie front, side (left or right), back and full body. Example images from the two datasets are demonstrated in Fig.~\ref{fig: problem-setup}. We can see that the view and scale variance of images from DeepFashion is much higher than those in MVC. The total number of images in DeepFashion is also much smaller than MVC. Both the high variance and the limited number of training samples bring great difficulties to multi-view image generation on DeepFashion. To give a consistent task specification of multi-view image generation on the two datasets, we define that the view set contains the front, side and back view. We consider two generation goals: to generate the side view and back view images conditioned on the front view image; and to generate the front view and back view image from side view images. These two scenarios are most popular in real life. We split the MVC dataset into the training set with 33,323 groups of images and the testing set with 3,000 groups of images. Each group contains the three views of clothing images. The training set of DeepFashion dataset consists of 7,897 groups of clothing images, and there are 800 groups of images in the testing set.

\vspace{-0.1in}
\paragraph{Evaluation Metrics}
In the previous literatures on image generation, the performance is usually evaluated by human subjects, which is subjective and time-consuming. Since the ground-truth images with target views are provided in the datasets, thus in our experiments, we adopt the Structural Similarity (SSIM)~\cite{Wang2004} to measure the similarity between generated image and ground truth image. We do not use the pixel-level mean square error as the evaluation metric since we focus more on the quality of generated clothes images for evaluation (note similarity of generated image and ground truth essentially also measures the quality of the result). There are usually human models in the generated clothes images, thus the images may present different poses (even in the same viewpoint) from the ground truth, which means pixel-wise mean square error is not suitable in our case.
SSIM can faithfully reflect the similarity of two images regardless of the light condition or small pose variance, since it models the perceived change in the structural information of the images. This evaluation metric is also widely used in many other image generation papers such as \cite{Yoo2016, Park2017}. The SSIM between two images $I_x$ and $I_y$ is defined as 
\begin{align}
  \text{SSIM}(I_x, I_y)=\frac{(2\mu_x \mu_y + c_1)(2\sigma_{xy}+c_2)}{(\mu^2_x+\mu^2_y+c_1)(\sigma^2_x+\sigma^2_y+c_2)},\nonumber
\end{align}
where $x$ and $y$ are the generated image and the ground-truth image. $\mu$ and $\sigma$ are the average and variance of the image. $c_1$ and $c_2$ are two variables to stabilize the division, which are determined by the dynamic range of the images.

The ``inception score''~\cite{Salimans2016} for quantitative evaluation is chosen as another metric in our paper as \cite{Zhang2016, Isola2016}. It is defined as
\begin{align}
  \text{IS}(I_x, y)=\exp(\mathbb{E}_{I_x} D_{\mathrm{KL}}(p(y|\bm{I_x})||p(y))),\nonumber
\end{align}
where $\bm{I_x}$ denotes one generated image, and $y$ is the label predicted by the Inception model. It is computed based on the assumption that the generated image with good quality should diverse and meaningful, \ie, the $\mathrm{KL}$ divergence between the marginal distribution $p(y)$ and the conditional distribution $p(y|I_x)$ should be large. 

\begin{figure*}[!t]
  \centering
  \subfloat[Input]{\includegraphics[height=1.8in]{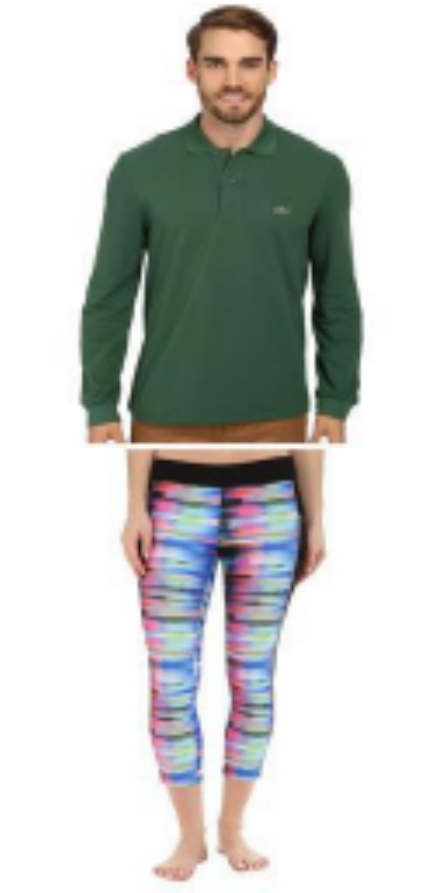}}
  \hspace{0.15in}
  \subfloat[cVAE]{\includegraphics[height=1.8in]{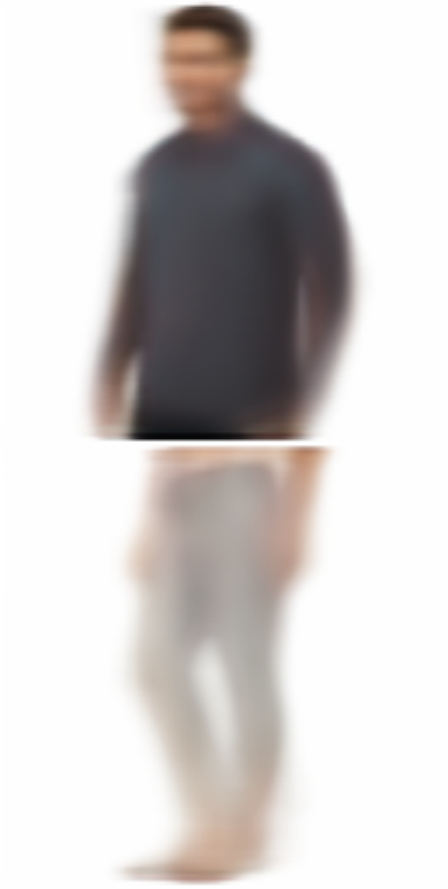}} 
  \hspace{0.15in}
  \subfloat[cGANs]{\includegraphics[height=1.8in]{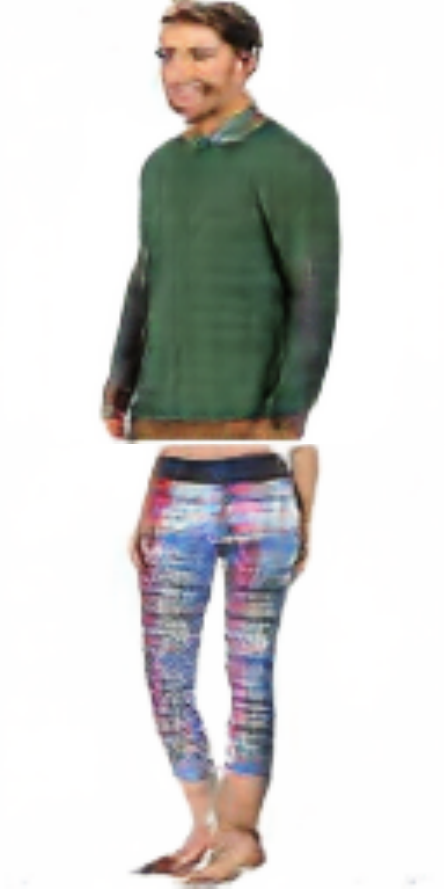}} 
  \hspace{0.15in}
  \subfloat[Ours]{\includegraphics[height=1.8in]{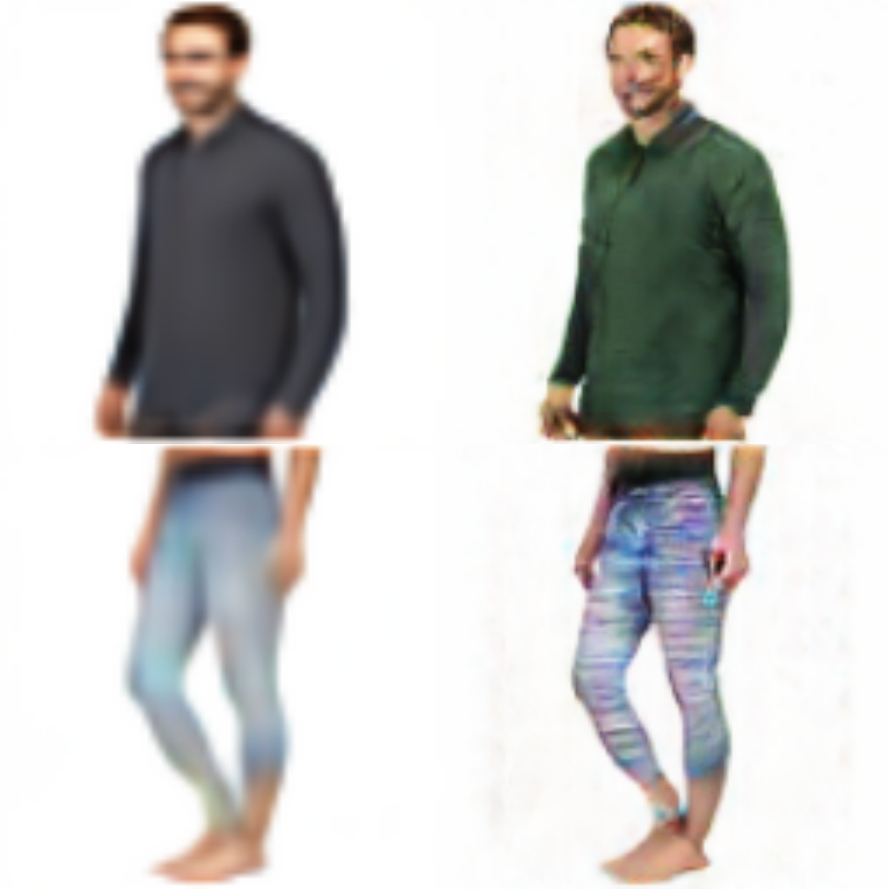}} 
  \hspace{0.15in}
  \subfloat[GT]{\includegraphics[height=1.8in]{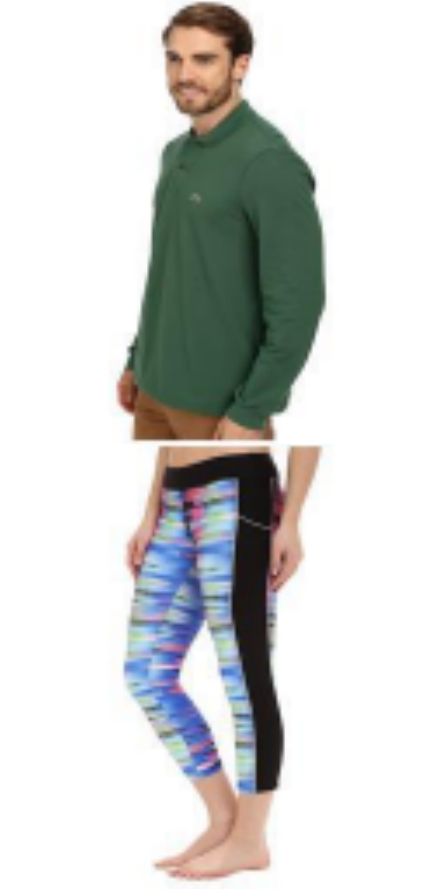}} 
  \caption{Example results by our proposed method and state-of-the-arts methods.}
  \label{fig: results-baselines}
  \vspace{-0.2in}
\end{figure*}

\begin{figure*}[!t]
  \centering
  \subfloat[Input Images]{\includegraphics[height=3.4in]{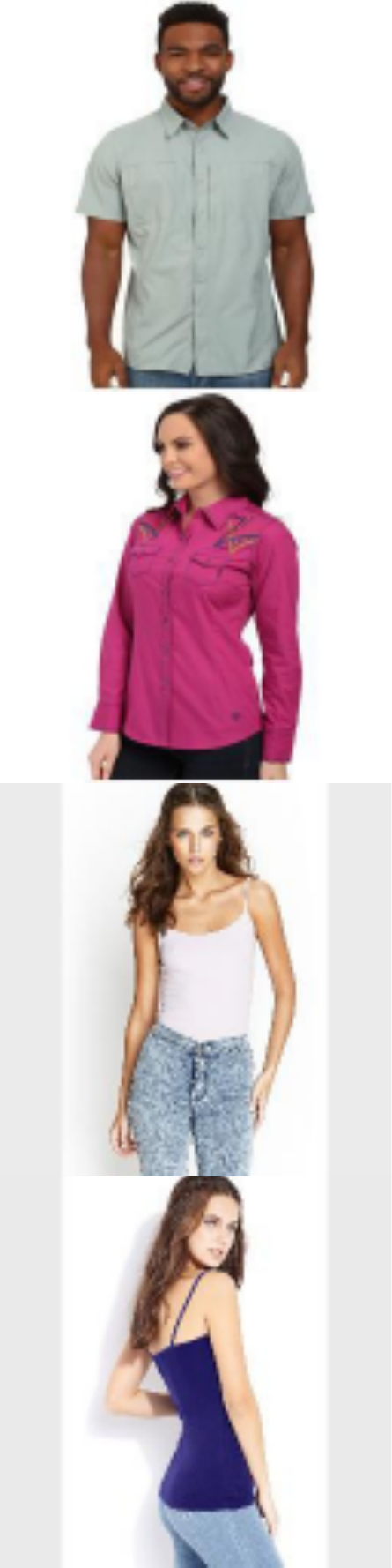}}
  \hspace{0.1in} 
  \subfloat[Coarse Images]{\includegraphics[height=3.4in]{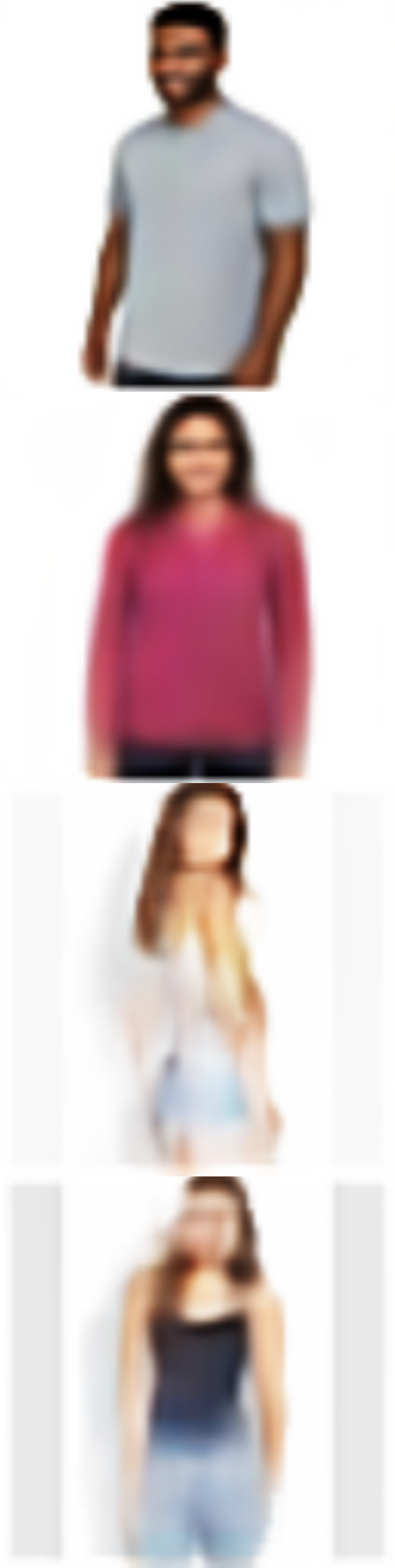}} 
  \hspace{0.1in} 
  \subfloat[Fine Images]{\includegraphics[height=3.4in]{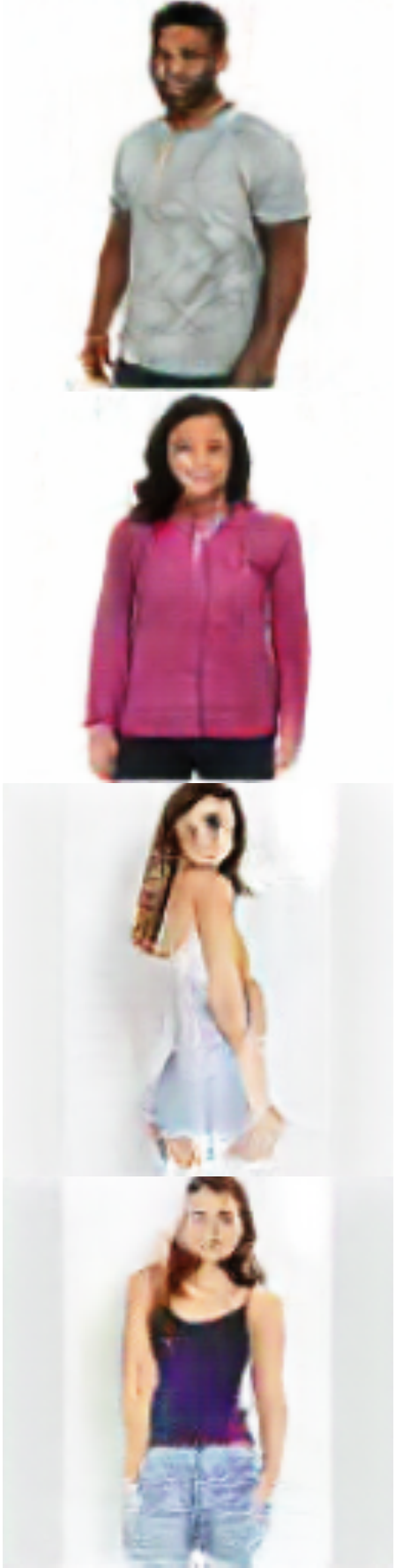}}
    \hspace{0.1in} 
  \subfloat[GT Images]{\includegraphics[height=3.4in]{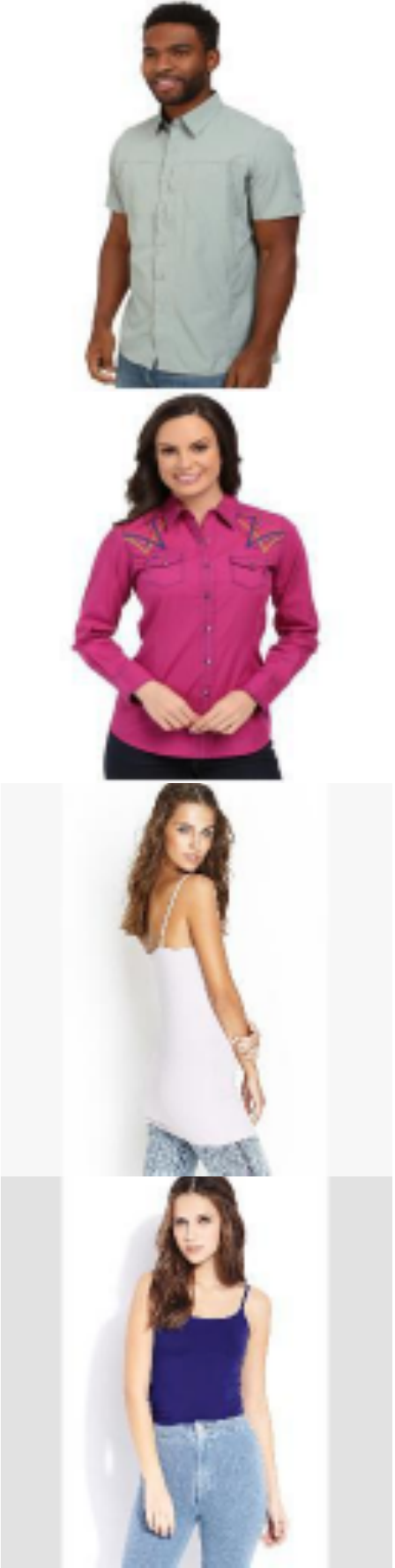}}
    \hspace{0.1in} 
  \subfloat[Coarse Images]{\includegraphics[height=3.4in]{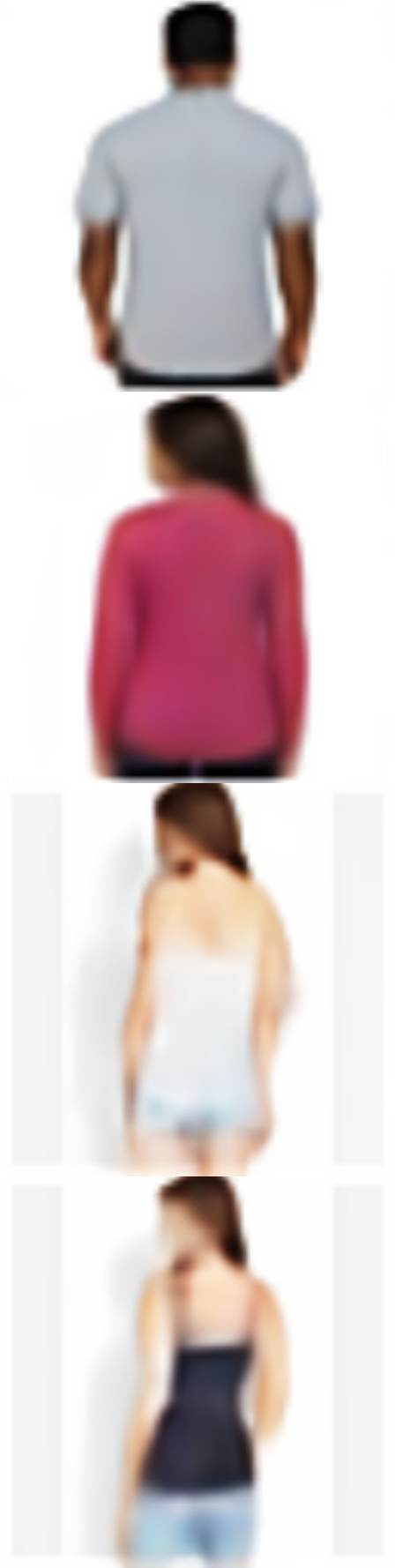}}
    \hspace{0.1in} 
  \subfloat[Fine Images]{\includegraphics[height=3.4in]{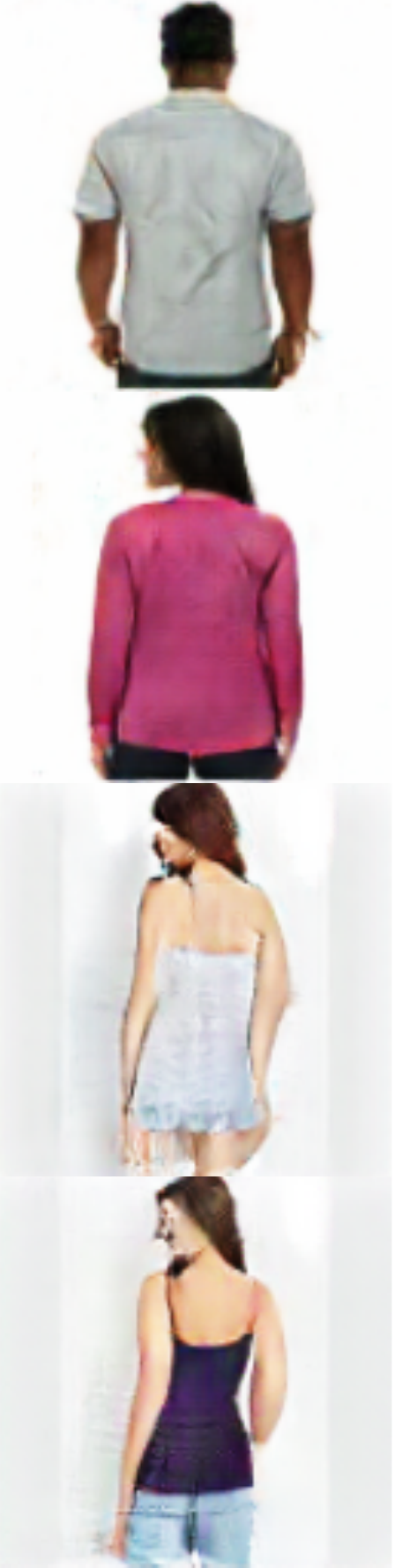}}
    \hspace{0.1in} 
  \subfloat[GT Images]{\includegraphics[height=3.4in]{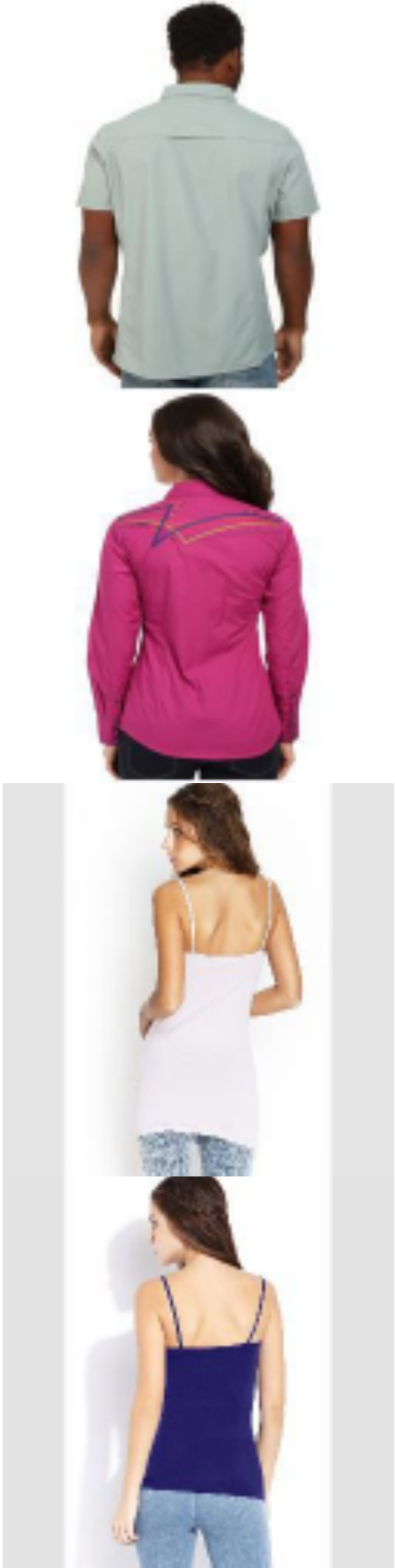}} 
  \vspace{0.1in}
  \caption{Example results by our proposed method. The first two rows shows examples from MVC, and the last two rows shows examples from DeepFashion. The images are generated from coarse to fine conditioned on input images of different views}
  \label{fig: results-vgan}
  \vspace{-0.2in}
\end{figure*}

\begin{figure*}[!t]
  \centering
  \subfloat[]{\includegraphics[height=1.1in]{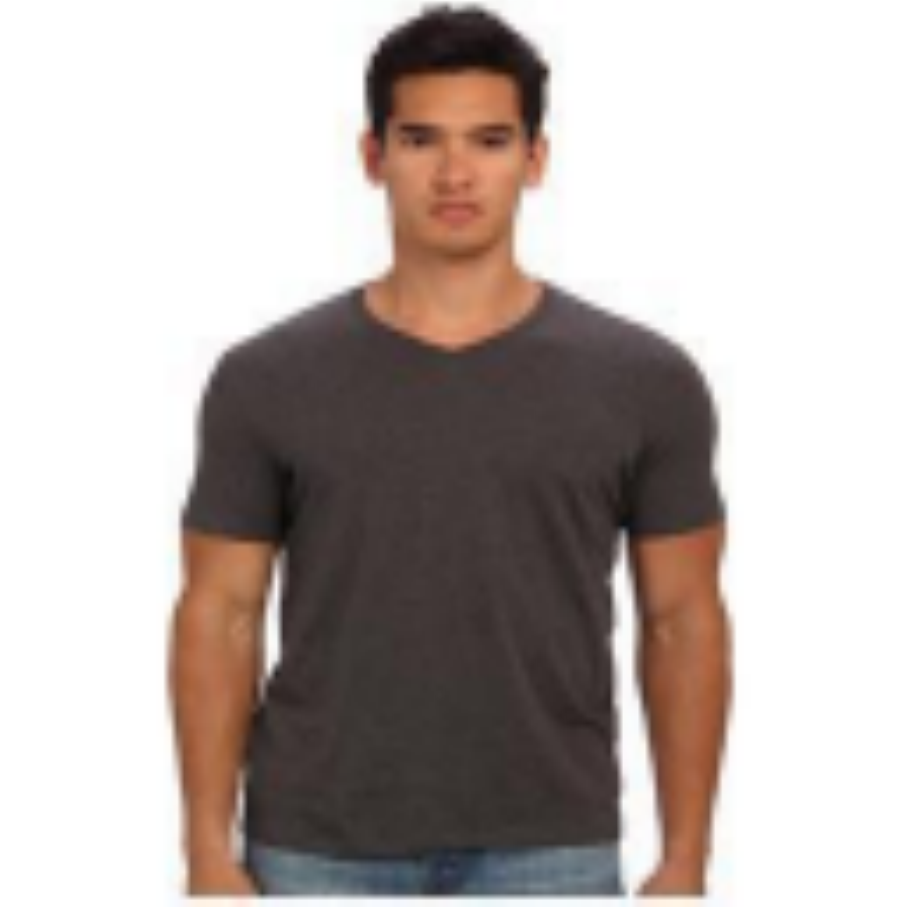}}
  \subfloat[]{\includegraphics[height=1.1in]{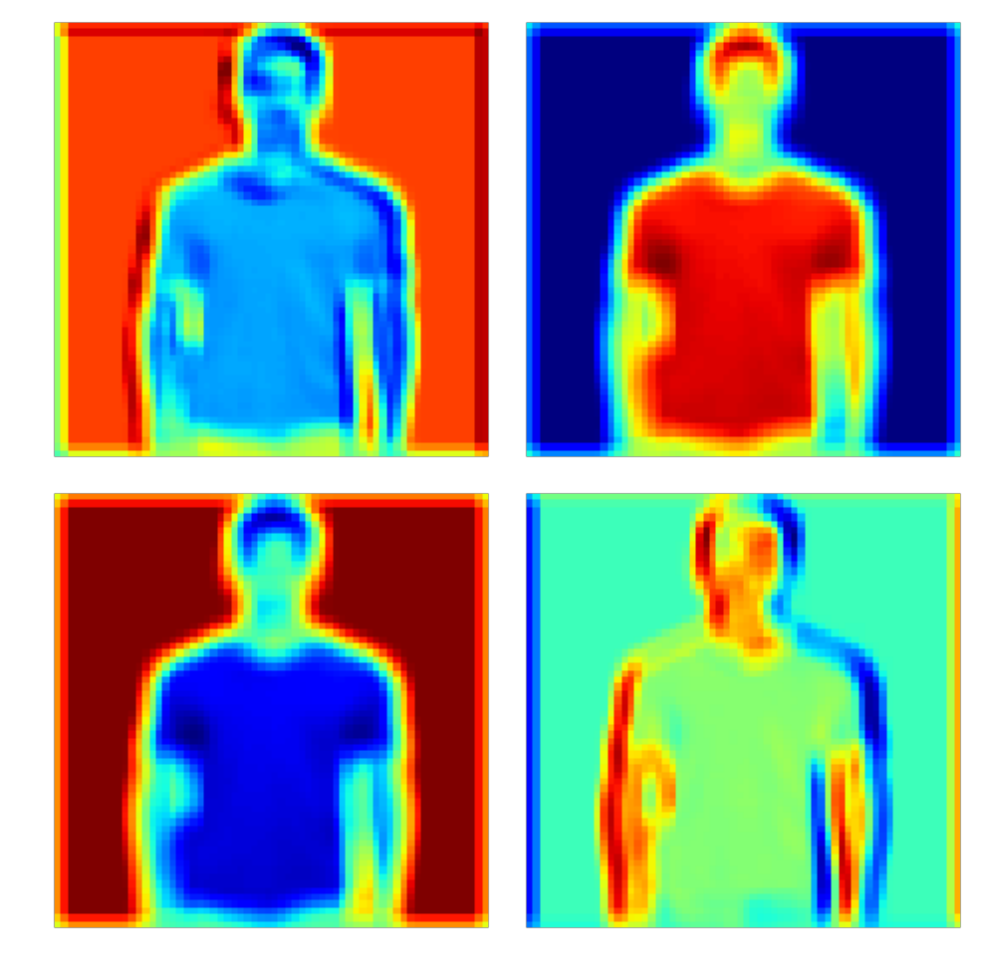}} 
  \hspace{0.05in}
  \subfloat[]{\includegraphics[height=1.1in]{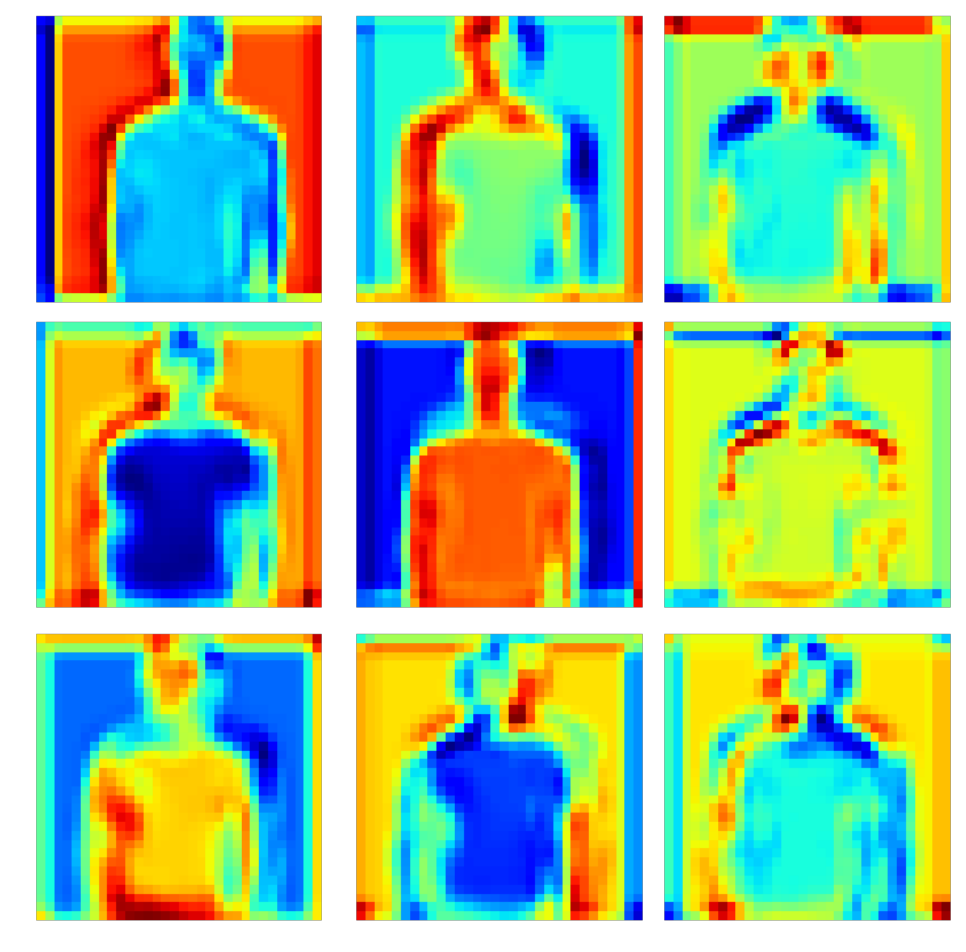}} 
  \hspace{0.05in}
  \subfloat[]{\includegraphics[height=1.1in]{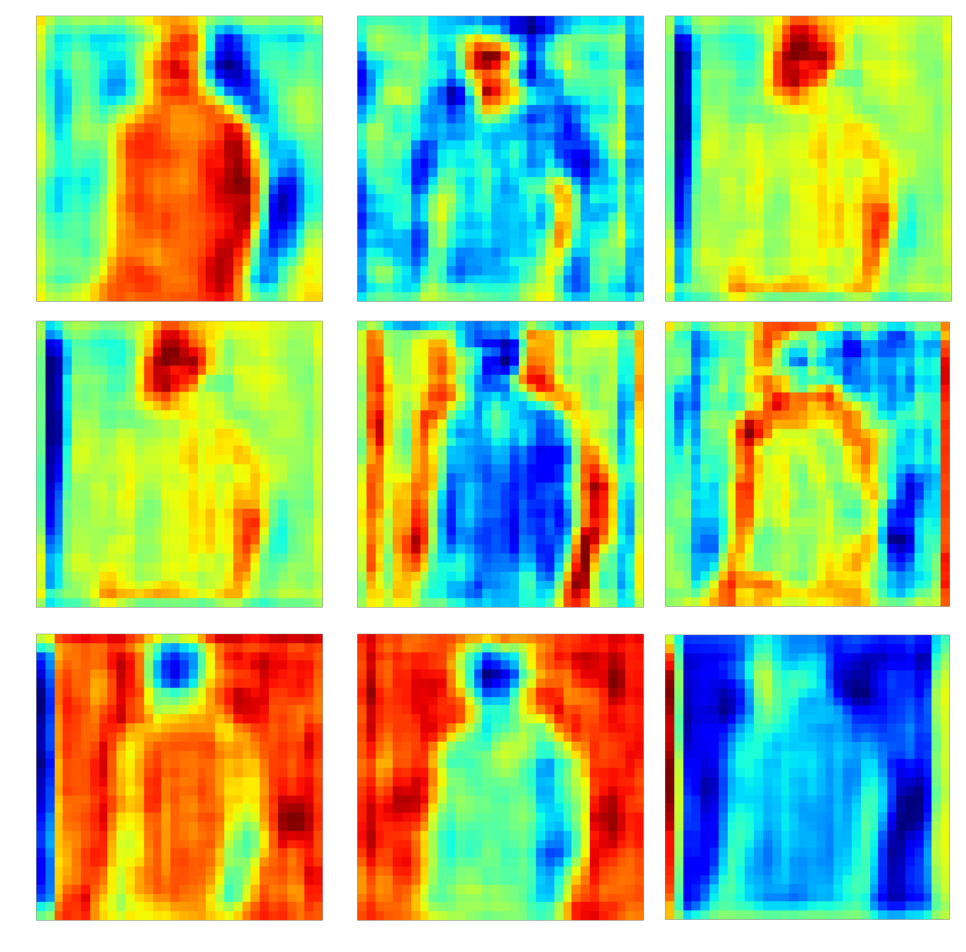}} 
  \hspace{0.05in}
  \subfloat[]{\includegraphics[height=1.1in]{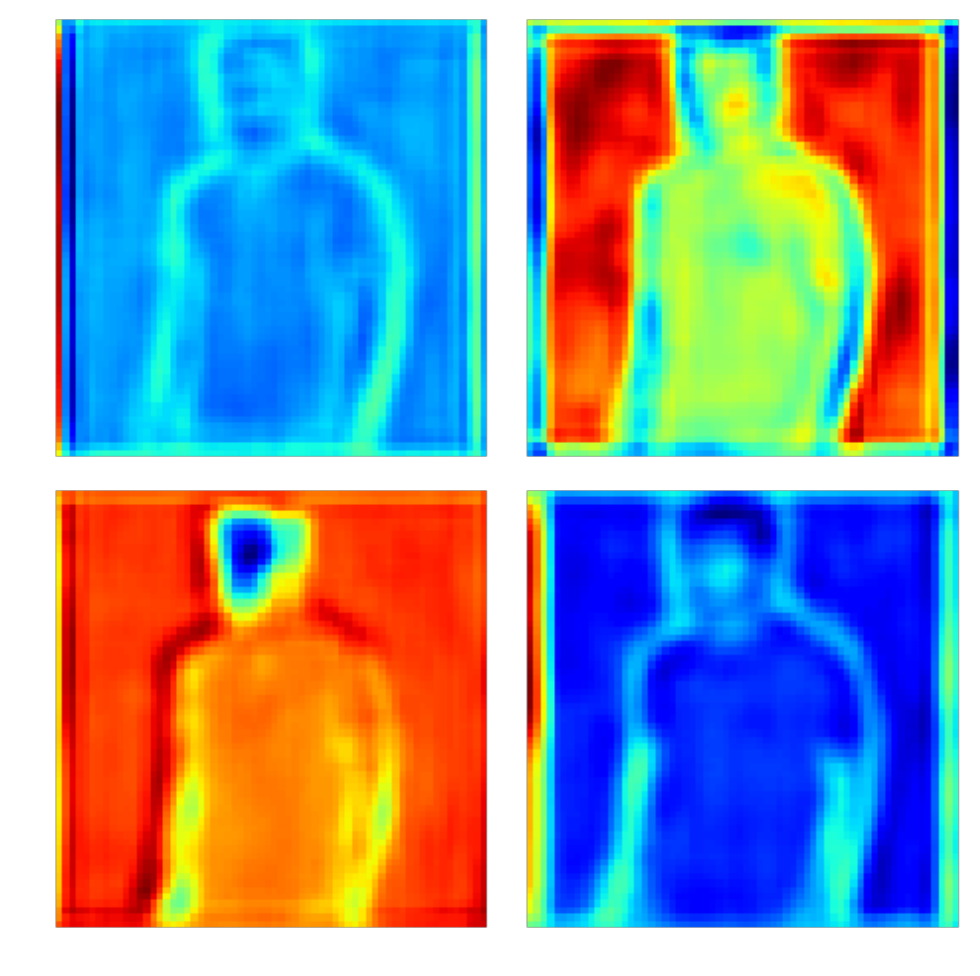}} 
  \subfloat[]{\includegraphics[height=1in]{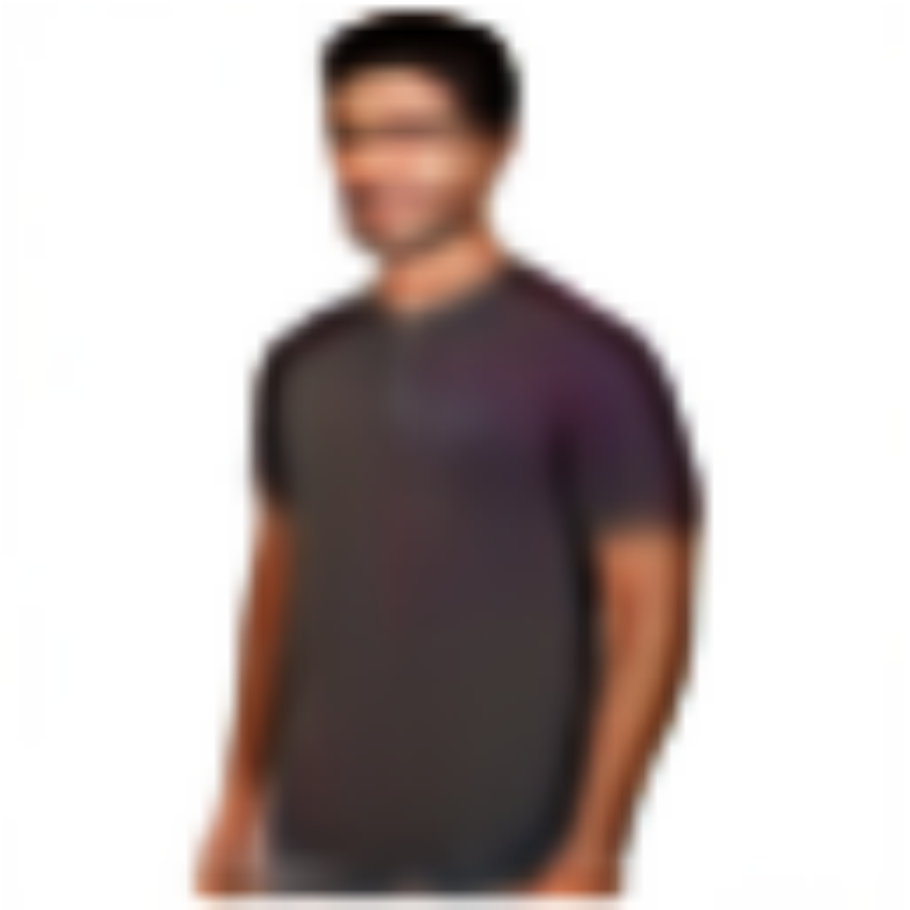}} 
  \caption{Visualization of feature maps in the first two convolution layers ((b) \& (c)) in the encoder of coarse image generation and the mirrored last two convolution layers ((d) \& (e)) in the decoder of coarse image generation. Our model learns how to transform image into the desired view. (a) and (f) are the input image and the generated LR image.}
  \label{fig: fm-vis}
  \vspace{-0.2in}
\end{figure*}

\begin{figure*}[!t]
  \centering
  \subfloat[]{\includegraphics[height=0.9in]{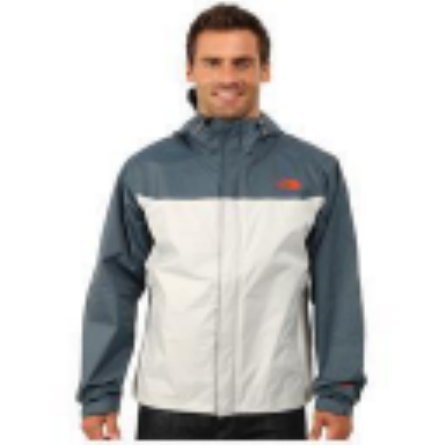}}
  \subfloat[]{\includegraphics[height=0.9in]{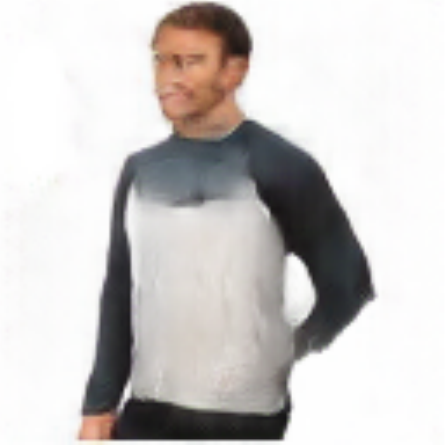}}
  \hspace{0.05in} 
  \subfloat[]{\includegraphics[height=0.9in]{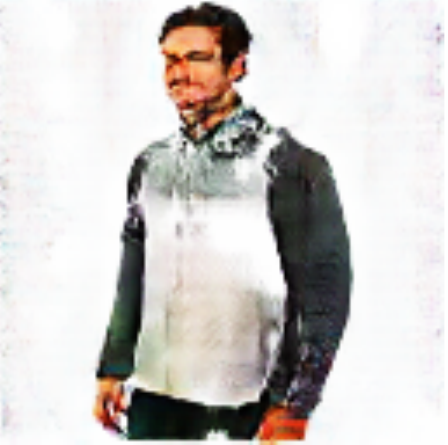}} 
  \hspace{0.05in} 
  \subfloat[]{\includegraphics[height=0.9in]{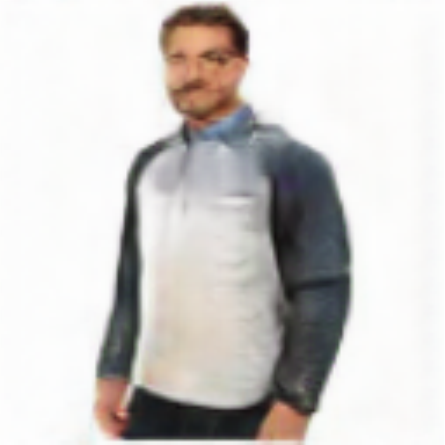}} 
  \hspace{0.05in}
  \subfloat[]{\includegraphics[height=0.9in]{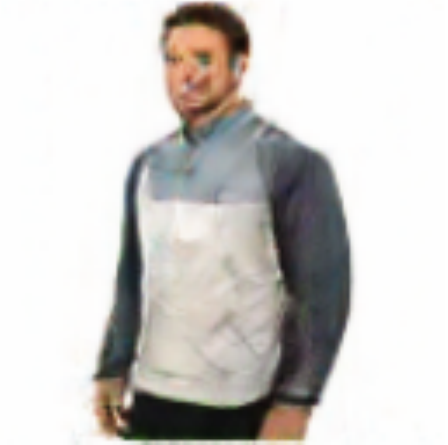}} 
  \hspace{0.05in} 
  \subfloat[]{\includegraphics[height=0.9in]{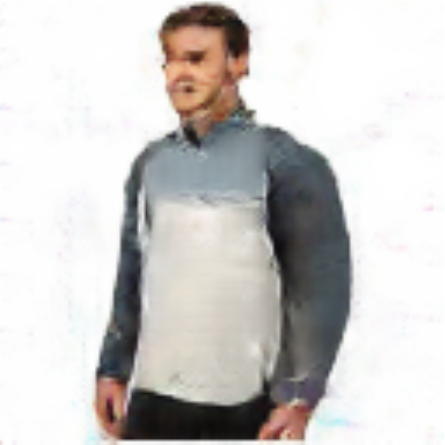}} 
  \hspace{0.05in} 
  \subfloat[]{\includegraphics[height=0.9in]{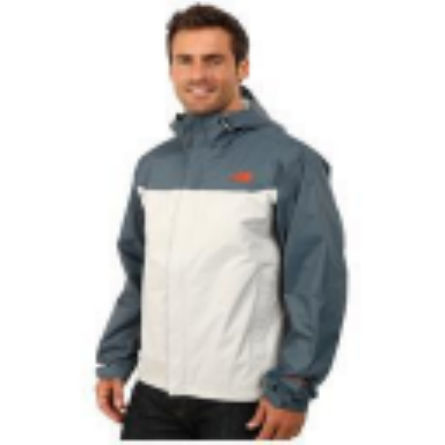}}
  \vspace{0.1in}
  \caption{Generated images with different variants of our proposed method. (b), (c), (d) and (e) are the results of the model without $V$, Dual-path U-Net, $\ell_1$ loss and conditional discriminator, respectively. (f) shows the results generated by our proposed model. (a) and (g) are the input images and the ground truth image.}
  \label{fig: results-ablation}
  \vspace{-0.1in}
\end{figure*}

\subsection{Experimental Results and Analysis}

\begin{table}[!b]
\footnotesize
\centering
\vspace{-0.1in}
\caption{SSIM and IS of our proposed method and state-of-the-arts methods.}
\vspace{0.1in}
\label{tb: comparison-baselines}
\begin{tabular}{|l|c|c|c|c|}
\hline
\multirow{2}{*}{Methods} & \multicolumn{2}{c|}{SSIM} & \multicolumn{2}{c|}{IS} \\ \cline{2-5} 
                  & MVC     & DF       & MVC    & DF  \\\hline
cVAE~\cite{Sohn2015} &  0.66 $\pm$ .09 & 0.58 $\pm$ .08 & 2.61 $\pm$ .06 & 2.35 $\pm$ .08 \\
cGANs~\cite{Mirza2014} & 0.69 $\pm$ .10 & 0.59 $\pm$ .08 & 3.45 $\pm$ .08 & 2.45 $\pm$ .10 \\
Ours  & \textbf{0.70} $\pm$ .10 & \textbf{0.62} $\pm$ .08 & \textbf{3.69} $\pm$ .09 & \textbf{3.03} $\pm$ .20\\ \hline
\end{tabular}
\end{table}

We compare our results with two state-of-the-arts methods, \ie Conditional VAE (cVAE)~\cite{Sohn2015} and Conditional GANs (cGANs)~\cite{Mirza2014} on MVC and DeepFashion datasets. The SSIM and Inception Scores for our proposed method and compared methods are reported in Table~\ref{tb: comparison-baselines}. We can see that cVAE has the worst SSIM and Inception Scores on both datasets, while the cGANs improve the SSIM and Inception Score compared to cVAE.
Our proposed VariGANs further improves the performance on both datasets. The better SSIM and Inception Scores also indicate that our proposed method is able to generate more realistic images conditioned on the single-view image and the target view.

We demonstrate some representative examples generated by the state-of-the-arts methods and our proposed method in Fig.~\ref{fig: results-baselines}. It can be seen that the samples generated by cVAE are blurry, and the color is not correct. However, it correctly draws the general shape of the person and the clothes in the side view.
The images generated by cGANs are more realistic with more details, but present severe artifacts. Some of the generated images look unrealistic, such as the example in the second row. 
The low resolution image generated by the proposed coarse image generator with our proposed VariGAN model presents better shape and contour than cVAE, benefiting from a more reasonable target setting for this phase (\textit{i.e.} only generating LR images). Besides, the generated LR image looks more natural than those generated by other baselines, in terms of the view point, shape and contour. Finally, the fine image generator fills correct color and adds richer and realistic texture to the LR image.

We give more examples generated by our VariGANs in Fig.~\ref{fig: results-vgan}. The first two rows are from MVC dataset, and the others are from the DeepFashion dataset. The first and third row show generating the side and back view images from the front view. Given the side view image, the second and fourth row demonstrate the generated front and back view images. There are coarse images, fine images and ground-truth images shown in each column of Fig.~\ref{fig: results-vgan} (a) and Fig.~\ref{fig: results-vgan} (b). It can be seen that the generated coarse images have the right view based on the conditioned image. The details are added to the coarse images reasonably by our fine image generation module. The results also demonstrate that the generated images need not be the same as the ground-truth image. There may be pose variance in the generated images like the generated front view image of the second example. Note that our proposed model focuses on clothes generation and does not consider humans in the image. Besides, some blocky artifacts can be observed in some examples, In the future, we would explore how to remove such artifacts by adopting more complicated models to  learn to generate sharper details. Nevertheless, the current results present sufficient details about novel views for users. 
 
\vspace{-0.1in}
\paragraph{Visualization of the Feature Maps.}
To  provide a deeper insight to the mechanism of the multi-view image generation in our proposed model, we also visualize the feature maps of the first two convolution layers in the encoder of coarse image generation and their corresponding de-convolution layers in the decoder (\ie the last two), as shown in Fig.~\ref{fig: fm-vis}. The visualization demonstrates that the model learns how to change the view of different parts of the image. From the visualization results, one can observe that the generated feature maps effectively capture the transition of view angles and the counters from another view. 

\subsection{Ablation Study}

\begin{table}[!tb]
\footnotesize
\centering
\caption{SSIM and IS of our proposed method and its variants.}
\label{tb: comparison-ablation}
\hspace{0.01in}
\begin{tabular}{|l|c|c|c|c|}
\hline
\multirow{2}{*}{Methods} & \multicolumn{2}{c|}{SSIM} & \multicolumn{2}{c|}{IS} \\ \cline{2-5} 
  & MVC     & DF    & MVC           & DF           \\\hline
Ours w/o V & 0.69 $\pm$ .11 & 0.59 $\pm$ .07 & 3.49 $\pm$ .08 & 2.72 $\pm$ .08\\
Ours w/o U-Net & 0.56 $\pm$ .08 & 0.53 $\pm$ .07 & 3.04 $\pm$ .06 & 2.38 $\pm$ .07\\
Ours w/o $\ell_1$ & 0.58 $\pm$ .09 & 0.49 $\pm$ .06 & 3.23 $\pm$ .08 & 2.47 $\pm$ .06\\ 
Ours w/o cDisc & 0.66 $\pm$ .09 & 0.55 $\pm$ .09 & 3.25 $\pm$ .15 & 2.56 $\pm$ .05\\ 
Ours & \textbf{0.70} $\pm$ .10 & \textbf{0.62} $\pm$ .08 & \textbf{3.69} $\pm$ .09 & \textbf{3.03} $\pm$ .20\\\hline
\end{tabular}
\vspace{-0.2in}
\end{table}

In this subsection, we analyze the effectiveness of the components in our proposed model on MVC and DeepFashion to further validate the design of our model by conducting following experiments:

\begin{enumerate}[label={(\arabic*)}]
  \item VariGANs w/o $V$. In this experiment, the variational inference is replaced by another GANs to investigate the role of variational inference in our proposed VariGANs.
  \vspace{-0.05in}
  \item VariGANs w/o U-Net. The LR image and the conditioned image go through the Siamese encoder in the fine image generator until a bottle-neck, and the outputs of the encoders are concatenated and fed into the decoder networks. 
  \vspace{-0.05in}
  \item VariGANs w/o $\ell_1$ loss. This experiment is to verify the importance of the traditional reconstruction loss in generating plausible images.
  \vspace{-0.05in}
  \item VariGANs w/o conditional discriminator. Only the generated HR images and ground truth images are sent to the discriminator separately.
\end{enumerate} 

We report the results of those experiments on MVC and DeepFashion in Table~\ref{tb: comparison-ablation}. It can be seen that removing or replacing any component of our model lowers the performance of SSIM and IS. We also illustrate the images generated by different variants of our VariGANs in Fig.~\ref{fig: results-ablation}. Conditioned on the LR image generated by GANs, the result in Fig.~\ref{fig: results-ablation}.(b) displays relative good shape and right texture. However, there are also missing parts, \ie the left hand, in the generated image. The results generated by VariGANs without the dual-path U-Net have incomplete areas and un-natural colors as shown in Fig.~\ref{fig: results-ablation}.(c). Without $\ell_1$ loss, the detail texture is not learned well such as the upper part of the cloth in Fig.~\ref{fig: results-ablation}.(d). VariGANs without conditional discriminator generate comparative results (Fig.~\ref{fig: results-ablation}.(e)) as VariGANs (Fig.~\ref{fig: results-ablation}.(f)) except some smears.

%

\section{Conclusion}

In this paper, we propose a Variational Generative Adversarial Networks~(VariGANs) for synthesizing realistic clothing images with different views as input image. The proposed method enhances the GANs with variational inference, which generate image from coarse to fine. Specifically, providing the input image with a certain view, the coarse image generator first generate the basic shape of the object with target view. Then the fine image generator fill the details into the coarse image and correct the defects.
With extensive experiments, our model can generate more plausible results than the state-of-the-arts methods. The ablation studies also verify the importance of each component in the proposed VariGANs.

{\small
\bibliographystyle{ieee}
\bibliography{iccv}

\begin{thebibliography}{10}\itemsep=-1pt

\bibitem{Chen2013}
T.~Chen, Z.~Zhu, A.~Shamir, S.-M. Hu, and D.~Cohen-Or.
\newblock 3-sweep: Extracting editable objects from a single photo.
\newblock {\em ACM Transactions on Graphics}, 2013.

\bibitem{Chen2016}
X.~Chen, Y.~Duan, R.~Houthooft, J.~Schulman, I.~Sutskever, and P.~Abbeel.
\newblock Infogan: Interpretable representation learning by information
  maximizing generative adversarial nets.
\newblock {\em arXiv:1606.03657}, 2016.

\bibitem{Donahue2016}
J.~Donahue, P.~Kr{\"a}henb{\"a}hl, and T.~Darrell.
\newblock Adversarial feature learning.
\newblock {\em arXiv:1605.09782}, 2016.

\bibitem{Dosovitskiy2015}
A.~Dosovitskiy, J.~T. Springenberg, M.~Tatarchenko, and T.~Brox.
\newblock Learning to generate chairs, tables and cars with convolutional
  networks.
\newblock In {\em CVPR}, 2015.

\bibitem{Dumoulin2017}
V.~Dumoulin, I.~Belghazi, B.~Poole, O.~Mastropietro, A.~Lamb, M.~Arjovsky, and
  A.~Courville.
\newblock Adversarially learned inference.
\newblock {\em arXiv:1606.00704}, 2017.

\bibitem{Goodfellow2014}
I.~Goodfellow, J.~Pouget-Abadie, M.~Mirza, B.~Xu, D.~Warde-Farley, S.~Ozair,
  A.~Courville, and Y.~Bengio.
\newblock Generative adversarial nets.
\newblock In {\em NIPS}, 2014.

\bibitem{Gregor2015}
K.~Gregor, I.~Danihelka, A.~Graves, D.~J. Rezende, and D.~Wierstra.
\newblock Draw: A recurrent neural network for image generation.
\newblock In {\em ICML}, 2015.

\bibitem{Hinton2011}
G.~E. Hinton, A.~Krizhevsky, and S.~D. Wang.
\newblock Transforming auto-encoders.
\newblock In {\em ICANN}, 2011.

\bibitem{Isola2016}
P.~Isola, J.-Y. Zhu, T.~Zhou, and A.~A. Efros.
\newblock Image-to-image translation with conditional adversarial networks.
\newblock {\em arXiv:1611.07004}, 2016.

\bibitem{Kholgade2014}
N.~Kholgade, T.~Simon, A.~Efros, and Y.~Sheikh.
\newblock 3d object manipulation in a single photograph using stock 3d models.
\newblock {\em ACM Transactions on Graphics}, 2014.

\bibitem{Kingma2014}
D.~P. Kingma and M.~Welling.
\newblock Auto-encoding variational bayes.
\newblock In {\em ICLR}, 2014.

\bibitem{Kulkarni2015}
T.~D. Kulkarni, W.~Whitney, P.~Kohli, and J.~B. Tenenbaum.
\newblock Deep convolutional inverse graphics network.
\newblock In {\em NIPS}, 2015.

\bibitem{Liu2016}
K.-H. Liu, T.-Y. Chen, and C.-S. Chen.
\newblock Mvc: A dataset for view-invariant clothing retrieval and attribute
  prediction.
\newblock In {\em ICMR}, 2016.

\bibitem{Liu2016a}
Z.~Liu, P.~Luo, S.~Qiu, X.~Wang, and X.~Tang.
\newblock Deepfashion: Powering robust clothes recognition and retrieval with
  rich annotations.
\newblock In {\em CVPR}, 2016.

\bibitem{Mirza2014}
M.~Mirza and S.~Osindero.
\newblock Conditional generative adversarial nets.
\newblock {\em arXiv:1411.1784}, 2014.

\bibitem{Odena2016}
A.~Odena.
\newblock Semi-supervised learning with generative adversarial networks.
\newblock {\em arXiv:1606.01583}, 2016.

\bibitem{Odena2016a}
A.~Odena, C.~Olah, and J.~Shlens.
\newblock Conditional image synthesis with auxiliary classifier gans.
\newblock {\em arXiv:1610.09585}, 2016.

\bibitem{Park2017}
E.~Park, J.~Yang, E.~Yumer, D.~Ceylan, and A.~C. Berg.
\newblock Transformation-grounded image generation network for novel 3d view
  synthesis.
\newblock In {\em CVPR}, 2017.

\bibitem{Pathak2016}
D.~Pathak, P.~Krahenbuhl, J.~Donahue, T.~Darrell, and A.~A. Efros.
\newblock Context encoders: Feature learning by inpainting.
\newblock In {\em CVPR}, 2016.

\bibitem{Reed2016}
S.~Reed, Z.~Akata, X.~Yan, L.~Logeswaran, B.~Schiele, and H.~Lee.
\newblock Generative adversarial text-to-image synthesis.
\newblock In {\em ICML}, 2016.

\bibitem{Rezende2016}
D.~J. Rezende, S.~M.~A. Eslami, S.~Mohamed, P.~W. Battaglia, M.~Jaderberg, and
  N.~Heess.
\newblock Unsupervised learning of 3d structure from images.
\newblock In {\em NIPS}, 2016.

\bibitem{Rezende2014}
D.~J. Rezende, S.~Mohamed, and D.~Wierstra.
\newblock Stochastic backpropagation and approximate inference in deep
  generative models.
\newblock In {\em ICML}, 2014.

\bibitem{Ronneberger2015}
O.~Ronneberger, P.~Fischer, and T.~Brox.
\newblock U-net: Convolutional networks for biomedical image segmentation.
\newblock In {\em MICCAI}, 2015.

\bibitem{Salimans2016}
T.~Salimans, I.~Goodfellow, W.~Zaremba, V.~Cheung, A.~Radford, and X.~Chen.
\newblock Improved techniques for training gans.
\newblock {\em arXiv:1606.03498}, 2016.

\bibitem{Sohn2015}
K.~Sohn, H.~Lee, and X.~Yan.
\newblock Learning structured output representation using deep conditional
  generative models.
\newblock In {\em NIPS}, 2015.

\bibitem{Wang2004}
Z.~Wang, A.~C. Bovik, H.~R. Sheikh, and E.~P. Simoncelli.
\newblock Image quality assessment: From error visibility to structural
  similarity.
\newblock {\em IEEE Transactions on Image Processing}, 13(4):600--612, 2004.

\bibitem{Wu2016a}
J.~Wu, T.~Xue, J.~J. Lim, Y.~Tian, J.~B. Tenenbaum, A.~Torralba, and W.~T.
  Freeman.
\newblock Single image 3d interpreter network.
\newblock In {\em ECCV}, 2016.

\bibitem{Yan2016}
X.~Yan, J.~Yang, K.~Sohn, and H.~Lee.
\newblock Attribute2image: Conditional image generation from visual attributes.
\newblock In {\em ECCV}, 2016.

\bibitem{Yan2016a}
X.~Yan, J.~Yang, E.~Yumer, Y.~Guo, and H.~Lee.
\newblock Perspective transformer nets: Learning single-view 3d object
  reconstruction without 3d supervision.
\newblock In {\em NIPS}, 2016.

\bibitem{Yoo2016}
D.~Yoo, N.~Kim, S.~Park, A.~S. Paek, and I.~S. Kweon.
\newblock Pixel-level domain transfer.
\newblock {\em arXiv:1603.07442}, 2016.

\bibitem{Zhang2016}
H.~Zhang, T.~Xu, H.~Li, S.~Zhang, X.~Huang, X.~Wang, and D.~Metaxas.
\newblock Stackgan: Text to photo-realistic image synthesis with stacked
  generative adversarial networks.
\newblock {\em arXiv:1612.03242}, 2016.

\bibitem{Zheng2012}
Y.~Zheng, X.~Chen, M.-M. Cheng, K.~Zhou, S.-M. Hu, and N.~J. Mitra.
\newblock Interactive images: cuboid proxies for smart image manipulation.
\newblock {\em ACM Transactions on Graphics}, 2012.

\bibitem{Zhou2016a}
T.~Zhou, S.~Tulsiani, W.~Sun, J.~Malik, and A.~A. Efros.
\newblock View synthesis by appearance flow.
\newblock In {\em ECCV}, 2016.

\bibitem{Zhou2016}
Y.~Zhou and T.~L. Berg.
\newblock Learning temporal transformations from time-lapse videos.
\newblock In {\em ECCV}, 2016.

\bibitem{Zhu2016}
J.-Y. Zhu, P.~Kr{\"a}henb{\"u}hl, E.~Shechtman, and A.~A. Efros.
\newblock Generative visual manipulation on the natural image manifold.
\newblock In {\em ECCV}, 2016.

\end{thebibliography}
}

\end{document}